\documentclass[letterpaper, 10 pt, conference]{ieeeconf}  % Comment this line out if you need a4paper

\IEEEoverridecommandlockouts                              % This command is only needed if 
\usepackage[english]{babel}
\usepackage[utf8]{inputenc}
\usepackage{microtype}

\usepackage{flushend}

\usepackage{amsmath,amsfonts,amssymb}
\usepackage{amsthm}
\newtheorem{problem}{Problem}
\newtheorem{definition}{Definition}

\usepackage[hidelinks]{hyperref}
\usepackage{booktabs}
\usepackage{subfig}
\usepackage{graphicx}
\usepackage{yaacro}

\title{\LARGE \bf
Minimal Exposure Dubins Orienteering Problem
}

\author{Author$^{1}$ and Researcher$^{2}$% <-this % stops a space
\thanks{$^{1}$Author is with {\tt\small albert.author@papercept.net}}%
\thanks{$^{2}$Researcher is with {\tt\small b.d.researcher@ieee.org}}%
}

\author{Douglas G. Macharet, Armando Alves Neto, and Daigo Shishika% <-this % stops a space

\thanks{Douglas G. Macharet is with the Computer Vision and Robotics Laboratory (VeRLab), Department of Computer Science, Universidade Federal de Minas Gerais, Brazil. E-mail: {\tt\small doug@dcc.ufmg.br}.}
\thanks{Armando Alves Neto is with the Department of Electronics Engineering, Universidade Federal de Minas Gerais, MG, Brazil. E-mail: {\tt\small aaneto@cpdee.ufmg.br}}
\thanks{Daigo Shishika is with the Mechanical Engineering Department, George Mason University, USA. E-mail: {\tt\small dshishik@gmu.edu}}%
}

\newcommand{\escalar}[1]{\ensuremath{\mathit{#1}}}
\newcommand{\vetor}[1]{\ensuremath{\vec{\lowercase{#1}}}} %\mathbf
\newcommand{\set}[1]{\ensuremath{\boldsymbol{\mathcal{#1}}}}

\newcommand{\xpos}{\vetor{x}}
\newcommand{\sfunc}{S}
\newcommand{\ifunc}{I}
\newcommand{\efunc}{E}

\newcommand{\Rset}{\ensuremath{\mathbb{R}}}
\newcommand{\SE}[1]{\text{SE}(#1)}

\newcommand{\linvel}{\escalar{v}}

\newcommand{\rhomin}{\ensuremath{\rho_{\textrm{min}}}}
\newcommand{\rhomax}{\ensuremath{\rho_{\textrm{max}}}}

\newcommand{\positionset}{\set{P}}
\newcommand{\position}[1]{\ensuremath{\vetor{p}_{#1}}}
\newcommand{\nwaypoints}{\escalar{m}}

\newcommand{\dubinscurve}[1][\rhomin]{\ensuremath{\mathcal{D}_{#1}}}
\newcommand{\dubinslength}[1][\rhomin]{\ensuremath{\escalar{d}_{#1}}}
\newcommand{\dubinscircuit}{\ensuremath{\mathcal{L}}}

\newcommand{\vehicleconf}[1]{\ensuremath{\vetor{q}_{#1}}}

\newcommand{\vehicleori}[1]{\ensuremath{\theta_{#1}}}
\newcommand{\reward}{\escalar{r}}

\newcommand{\nodeset}{\set{N}}
\newcommand{\node}{\vetor{n}}
\newcommand{\nnodes}{\escalar{n}}

\newcommand{\sequence}[1]{\ensuremath{\langle #1 \rangle}}
\newcommand{\permutation}{\ensuremath{\Sigma}}

\newcommand{\budget}{\ensuremath{T_\textrm{max}}}

\newcommand{\sensingGain}{\escalar{\alpha}}

\usepackage[colorinlistoftodos]{todonotes}

\begin{acgroupdef}[list=acronimos]
    \acdef{TSP}{Traveling Salesman Problem}
    \acdef{DTSP}{Dubins Traveling Salesman Problem}
    \acdef{DTSPN}{Dubins Traveling Salesman Problem with Neighborhoods}
    
    \acdef{VNS}{Variable Neighborhood Search}
    
    \acdef{GA}{Genetic Algorithm}
  
    \acdef{OP}{Orienteering Problem}
    \acdef{DOP}{Dubins Orienteering Problem}
    
    \acdef{MEP}{Minimal Exposure Path}
  
    \acdef{WSN}{Wireless Sensor Network}
    \acdef[variantof=WSN]{WSNs}{Wireless Sensor Networks}
  
    \acdef{UAV}{Unmanned Aerial Vehicle}
    \acdef[variantof=UAV]{UAVs}{Unmanned Aerial Vehicles}
  
    \acdef{MEDOP}{Minimal Exposure Dubins Orienteering Problem}
\end{acgroupdef}

\begin{document}

\maketitle
\thispagestyle{empty}
\pagestyle{empty}

%\aaneto{Title suggestions: \emph{"Minimal Exposure Orienteering Problem for Dubins Vehicles"} or \emph{"Dubins Minimal Exposure Orienteering Problem", or "Dubins Orienteering Problem with Minimal Exposure"}}

%%%%%%%%%%%%%%%%%%%%%%%%%%%%%%%%%%%%%%%%%%%%%%%%%%%%%%%%%%%%%%

\begin{abstract}

    Different applications, such as environmental monitoring and military operations, demand the observation of predefined target locations, and an autonomous mobile robot can assist in these tasks.
    In this context, the \ac{OP} is a well-known routing problem, in which the goal is to maximize the objective function by visiting the most rewarding locations, however, respecting a limited travel budget (e.g., length, time, energy).
    %
    %The \ac{OP} is a significant challenge in the context of robotic exploration and surveillance, which consists of maximizing data collected from specific way-points by an agent subject to a limited traveling budget. Many papers in the literature have addressed this NP-hard routing problem, including scenarios where the agent is subject to bounded curvatures.
    %
    However, traditional formulations for routing problems generally neglect some environment peculiarities, such as obstacles or threatening zones.
    In this paper, we tackle the \ac{OP} considering Dubins vehicles in the presence of a known deployed sensor field.
    %
    %In this more realistic scenario, there are two main objectives: (i) maximize the collected reward, and (ii) minimize the exposure of the agent, i.e., the probability of being detected.
    %
    %However, classic \ac{OP} methods generally neglect environment constraints, such as obstacles or \emph{threatening zones}. Therefore, in this paper, we focus on solving the \ac{OP} for Dubins vehicles in two-dimensional spaces cluttered with \emph{detecting nodes}. In this more realistic scenario, there are two objectives: (i) \emph{maximize the collected reward} and (ii) \emph{minimize the threatening exposure} of the robot to a known sensor field detection given a limited budget.
    %
    We propose a novel multi-objective formulation called \ac{MEDOP}, whose main objectives are: (i) maximize the collected reward, and (ii) minimize the exposure of the agent, i.e., the probability of being detected.
    The solution is based on an evolutionary algorithm that iteratively varies the subset and sequence of locations to be visited, the orientations on each location, and the turning radius used to determine the paths. Results show that our approach can efficiently find a diverse set of solutions that simultaneously optimize both objectives.
    
    %Results show a compromise solution between both objectives.
    %
    %
    %In the context of robotic exploration and surveillance, recent papers have concentrated on two interesting and distinct problems: the \ac{MEP} and the \ac{OP}. The first one consists of finding paths that less exposure the robot to threatening zones in the environments, while the second one focuses on maximizing data collected from specific way-points subject to a limited traveling budget.
    %
    % .
    %We introduce and formulate this new problem.
    %In this paper, however, we propose a planning algorithm to deal simultaneously with both problems. Our focus is on information gathering using curvature-constrained (Dubins) vehicles, whose objectives are \emph{maximize the collected reward} and \emph{minimize the threatening exposure} of the robot to a known sensor field detection, given a limited budget.
    %
    %We present a novel multi-objective formulation called \ac{MEDOP}, whose solution is based on an evolutionary algorithm that iteratively varies the sequence of points to be visited, their respective orientations and the turning radius used to determine the paths.
\end{abstract}

%%%%%%%%%%%%%%%%%%%%%%%%%%%%%%%%%%%%%%%%%%%%%%%%%%%%%%%%%%%%%%
%%%%%%%%%%%%%%%%%%%%%%%%%%%%%%%%%%%%%%%%%%%%%%%%%%%%%%%%%%%%%%
\section{Introduction}

Advances on the research of autonomous vehicles have spread their use in applications such as environmental monitoring, surveillance, and military operations. In such cases, the tasks include typical \emph{routing problems}, whose solutions are variants of the classic \ac{TSP}. However, characteristics such as motion constraints and limited travel budget uncovers a broad range of new challenges to such well-known scenarios \cite{Macharet2018Survey}. 

%Endow a robot with the ability to collect information from an environment composed of propitious and threatening zones is a challenging issue.

%Thus, the \ac{TSP} has already been generalized in many ways to account for issues found in real-world applications \cite{Macharet2018Survey}. 
%
%The \ac{DTSP} is an example case, in which the result is circuit among visiting points taking into account the curvature constraint of Dubins vehicles \cite{Savla2005Point}. In order words, the route is feasible by robots with a limited turning radius, such as Ackerman (car-like) platforms or fixed-wing \ac{UAVs}.

The \ace{OP} is a variant of the \ac{TSP} which considers heterogeneous target locations to be visited, each with an associated reward, and a limited travel budget \cite{Chao1996AProblem}. The goal is to collect as many rewards as possible, respecting the budget. The \ac{OP} has been generalized to consider motion constraints \cite{Penicka2017DOP} and correlated rewards \cite{Tsiogkas2018DCOP}.
%
%In this context, two relevant problems have been treated in the literature: the \ace{OP} and the \ace{MEP}. In the \ac{OP}, the objective is to maximize the total reward collected in a subset of target locations, whose path length is limited by a given travel budget \cite{Chao1996AProblem}.
%
%On the other hand, in the \ac{MEP}, the goal is to design a path between two locations that minimize the exposure of the vehicle to local disturbances or to detection devices \cite{Veltri2003Minimal}.
%
%Although treated separately, these topics have much in common. Both are routing problems with high computational complexity, whose solutions must consider real-world issues such as motion constraints, obstacles, and uncertainties. In fact, \ac{OP} and \ac{MEP} are almost dual problems since in one there are regions to be visited, while in the other there are zones to be avoided.
%    
%Considering heterogeneous target locations to be visited, certain individual properties might result in different assigned \emph{rewards}, which can be seen as the location's importance. The \ac{OP} \cite{Chao1996AProblem}, considers the challenge of generating a path that maximizes the number of collected rewards, however, respecting a limited budget.
%
% Minimal Exposure / Military
However, traditional routing formulations alone may be insufficient to deal with more realistic scenarios. For example, surveillance and military applications might demand a path to have some particular properties, e.g., avoid certain zones in the environment. In this context, the \ac{MEP} problem \cite{Veltri2003Minimal} is relevant, whose goal is to design a path that traverses the environments and minimizes the probability of the vehicle being detected, or more generally, reduces the \emph{exposure} to any sort of threatening aspect.

%{
%\color{blue}
%In recent years there has been a growing awareness of the need for automated and assistive decision-making systems to move beyond single-objective formulations when dealing with complex real-world issues, which invariably involve multiple competing objectives. 
%}

\begin{figure}[t]
    \centering
    \includegraphics[width=\linewidth]{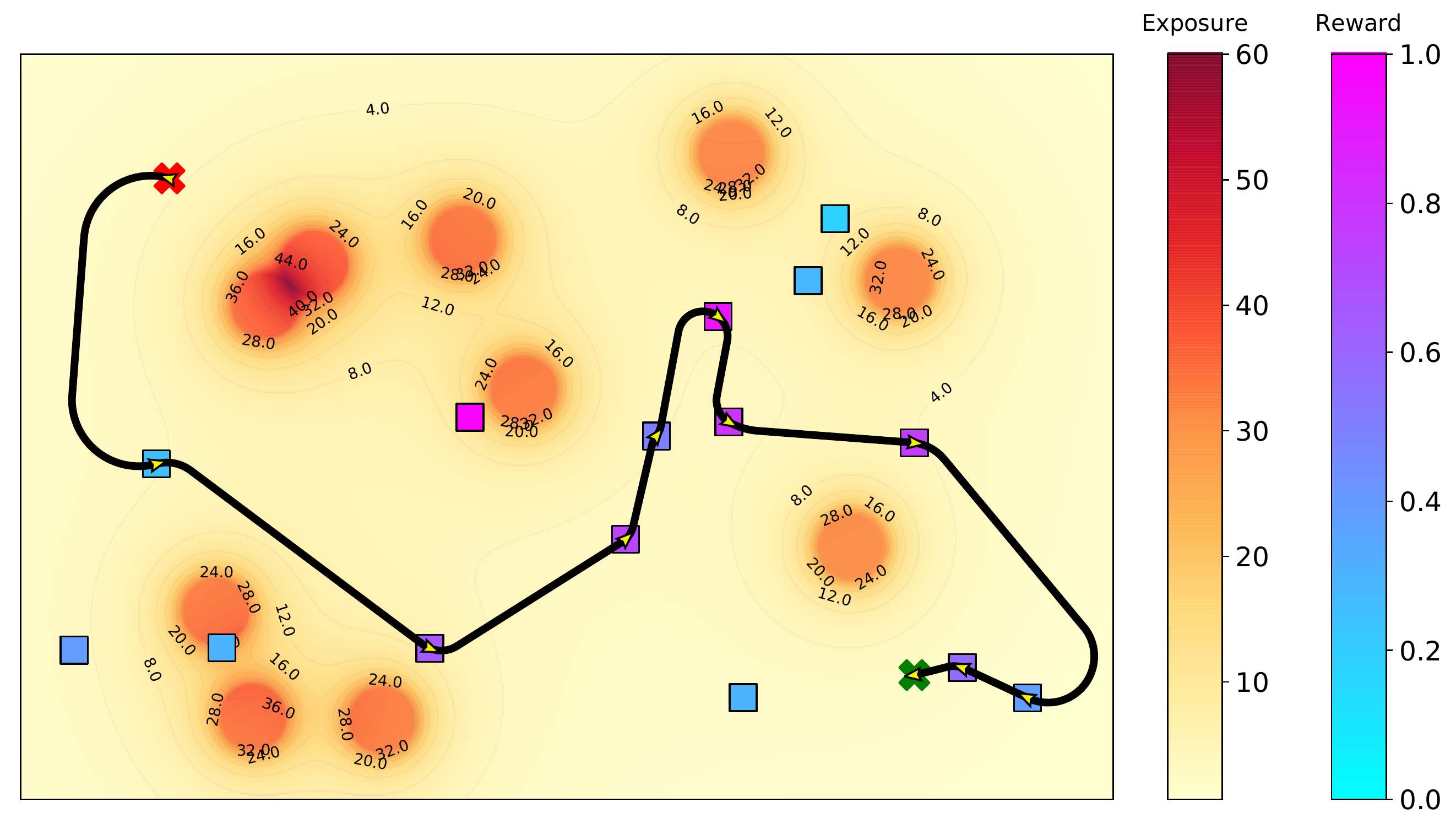}
    \caption{\protect\acd{MEDOP}: given initial and final positions (red/green Xs), the goal is to compute a curvature-constrained path (black line) that visits the most rewarding locations (colored squares). The path is subjected to a limited travel budget and it must minimize the exposure to a known sensor field (reddish regions).}
    %\caption{A Dubins vehicle must go from an initial position (red X) to a goal position (green X), visit the most rewarding subset of target locations (colored squares). The path (blue) should respect a travel budget and avoid detection by a sensor field in the environment (reddish surface).}
    % E: 2908.60 / R: 5.59 / L: 81.92
    \label{fig:example}
\end{figure}

%\douglas{TODO: Detail Problem.}
In this work, we tackle both of the aforementioned challenges, i.e., an \ac{OP} instance for vehicles with a bounded turning radius (Dubins vehicles) in an environment with a known deployed sensor field. The goal is to determine a path that is feasible by the vehicle and also respects a limited travel budget, maximizes the collected reward, and minimizes the exposure of the agent.
We combine these characteristics (motion constraints, limited travel budget, and exposure) into a novel unified multi-objective formulation, called \ace{MEDOP}. Figure \ref{fig:example} illustrates the problem, where a curvature-constrained path with varying turning radius is defined to avoid certain zones in the environment whilst visiting the most rewarding locations.

%Therefore, in this paper, we address the \ac{OP} into environments filled with detecting nodes, seeking to compute paths that maximize the collected reward in a subset of given locations while minimizing the exposure of the agent, similarly to the \ac{MEP} problem. Our robot is modeled as a Dubins platform, which increases complexity while makes its application more realistic.
%
%We define this combination of challenges (motion constraints, limited travel budget, and exposure) as a novel unified multi-objective formulation, called \ace{MEDOP}, whose an example is illustrated in Fig.~\ref{fig:example}. It shows a run where a curvature-constrained path (black line) is defined to avoid certain regions (red) in the environment whilst visiting the most rewarding locations.

%\douglas{TODO: Detail Solution.}
To solve this problem, we propose the use of an evolutionary algorithm that iteratively selects and evaluates distinct subsets of locations and in which order they should be visited; varies the orientations on each location; and attempts different feasible turning radii on the connecting segments, which aid to reduce the exposure.
Results confirm that there is an inverse relationship between the collected reward and the path exposure. In other words, to avoid certain zones the agent will probably have to visit a smaller number of locations. However, by analyzing the behavior of the Pareto front we conclude that the approach can efficiently find a diverse set of solutions that simultaneously optimize both objectives.

%To solve this problem, we propose an evolutionary algorithm that varies the sequence of visiting points, orientations and curvature radius, in order to find solutions that simultaneously optimize both objectives.
%
%Results show that there is an inverse relationship between reward and exposure, such that longer paths increase that amount of collected data, while let the robot be more exposed to detecting nodes. On the contrary, short paths present less exposure, but the reward also decreases.

%In the next section, we present the related work concerning both, \ac{OP} and \ac{MEP} approaches. Next, in Sec.~\ref{sec:preliminaries}, we provide the problem formalization. In Sec.~\ref{sec:methodology}, we present our proposed methodology, while the results are discussed in Sec.~\ref{sec:experiments}. Finally, in Sec.~\ref{sec:conclusion}, we present the conclusion and future work.

%%%%%%%%%%%%%%%%%%%%%%%%%%%%%%%%%%%%%%%%%%%%%%%%%%%%%%%%%%%%%%
%%%%%%%%%%%%%%%%%%%%%%%%%%%%%%%%%%%%%%%%%%%%%%%%%%%%%%%%%%%%%%
\section{Related Work}
\label{sec:related_work}

%OP: \cite{Chao1996AProblem}

A fundamental problem in Robotics is the generation of paths that are either length or time optimized, and a diverse collection of formulations and techniques to this challenge have been proposed in the literature.

% % The OP can be seen as the combination of the Knapsack  Problem  [4], when determining the subset of points to be visited,  and the  Traveling  Salesman  Problem[5], choosing the order in which these points will be visited.

The \ace{OP} \cite{Chao1996AProblem} is a multi-level optimization problem usually described as the combination of the Knapsack Problem (select subset of points) and the Traveling  Salesman  Problem (define the order of visit). The goal is to maximize the sum of the rewards associated with each location, however, considering a certain limited travel budget (e.g., length, time, energy, etc.). In most cases, this problem considers only a simple Euclidean metric to determine and represent the path, which may not be very representative.

%Therefore, it is necessary to determine both the subset of locations to be visited as well as the the order in which they will be visited.

%It consists of determining a path among a set points with associated rewards that maximizes the total reward. Such a path starts and ends in specific locations, and its length is bounded by a fixed traveling budget. Several solutions have been proposed to solve this NP-hard problem, also known as Euclidean \acd{OP}, since many applications can be mapped into it. However, all of them have neglected motion restrictions.

%D-OP: \cite{Penicka2017DOP,Vana2017Data,Penicka2017DOPN}

In real-world scenarios, a large number of vehicles present motion constraints, and such characteristics must be taken into account when searching for solutions \cite{LaValle2006Planning}. %The \ac{DTSP} \cite{Savla2005Point,Ny2012Dubins}, for example, is a \ac{TSP}-based problem in which the circuit is formed by Dubins curves \cite{Dubins1957Curves}.
Recently, the \ace{DOP} has been introduced \cite{Penicka2017DOP}, which considers the case of vehicles subjected to a bounded turning radius (Dubins vehicles \cite{Dubins1957Curves}). Besides the selection of the subset of locations and sequence of visit, in this case, the orientation on each position must also be determined, which increases the complexity. The authors have employed the \ac{VNS} metaheuristic to solve this NP-hard problem.
%
%Recently, variants of the \ac{OP} also considering Dubins vehicles have been addressed in two-dimensional \cite{Penicka2017DOP} and three-dimensional spaces \cite{Vana2017Data}.
%
%In the \ace{DOP}, Dubins curves were incorporated into the cost function to deal with the curvature constraint of some vehicles. Similarly to the classic \ac{DTSP}, nonholonomic characteristics required the definition of orientations at each point to be visited, which increases the complexity of the optimization routine.
%
%The approach has also been extended to visiting regions \cite{Penicka2017DOPN}, in which, similarly to the \ac{DTSPN}, subsections of the environment have been used instead of points to assign the visit reward.
%
%For all these \ac{DOP} variations, authors employed \ac{VNS} meta-heuristics to approximate the optimal solution.

%\ac{OP} and \ac{DOP} represent very complex NP-hard problems, but from a practical point-of-view, their solutions apply only for free navigation scenarios. In the real world, however, an agent may have to deal with obstacles or \emph{threatening zones} (areas with negative reward) while navigates towards positive rewarding points.
%
%That is similar to the \ace{MEP} problem.

Generally, routing problems such as the \ac{OP} and \ac{DOP} do not consider certain characteristics of the environment, and the determination of a safe path is also a fundamental requirement. Therefore, it is important to consider the presence of \emph{threatening zones}, which can encompass from regular obstacles to regions that have a more \emph{continuous} impact on the agent (e.g. detection probability).

The \ace{MEP} \cite{Veltri2003Minimal} problem considers the case where an agent must minimize the \emph{exposure} to certain regions in the environment. This concept introduced in \cite{Meguerdichian2001Exposure} defines exposure as the capacity a sensor field possesses of perceiving a target along an arbitrary path. This problem is critical as it is usually associated with the coverage quality of a \ac{WSN}.

%In the \ac{MEP}, a set of \ac{WSNs} are distributed inside a region of interest to detect moving targets. The \emph{exposure} of this target is given by the capacity the nodes possess of perceiving the target along the path. Then, from the robot's perspective, the goal is to plan a path across the environment that minimizes its exposure.
%
%This concept was introduced by \cite{Meguerdichian2001Exposure} and it is associated with the detection quality of the sensor field.%, represented by the exposure path with the minimum detection likelihood.

%MEP:
The first studies addressed the \ac{MEP} problem with classic methods, such as grid-based approaches and Voronoi diagrams \cite{Phipatanasuphorn2004Vulnerability, Djidjev2007Efficient}. Although simple to implement, such methods are not accurate enough, presenting difficulties in reaching the optimal solution \cite{Binh2019Efficient}. They also have limitations dealing with heterogeneous networks or a large number of sensor nodes \cite{Ye2016Hybrid}.
%
%\cite{Ferrari2010Potential} % obstaculos, multirrobo, campos potenciais
%\cite{Liu2014Minimal} % MEP com directional sensors
%\cite{Ye2016Hybrid}
%\cite{Binh2016Heuristic} % barrier coverage, voronoi
%
Due to the high complexity of the \ac{MEP}, most of the recent literature has employed heuristics such as genetic algorithms \cite{Feng2016Novel, Nguyen2017Genetic, Binh2019Efficient}.
%
%\cite{Feng2016Novel} % MEP, algoritmo genetico, voronoi
%\cite{Nguyen2017Genetic} % algoritmo genetico
%\cite{Binh2019Efficient} % algoritmo genetico
%
%\cite{Miller20113D} % 3D dubins, exposure em 2D (threats relief)
%\cite{Ozalp2013Optimal} % 3d, threatening zones, Algoritmo genetico
%\cite{Qu2014Optimal} %3D, Djikstra, campos potenciais
Different approaches have also been applied for aerial robots with dynamic constraints to avoid threatening zones in 3D \cite{Miller20113D, Qu2014Optimal}.

%\ac{OP} and \ac{MEP} are closely related problems. While in the \ac{OP} the environment is composed of rewarding points, whose amount of score must be maximized against a given budget, in the \ac{MEP} there is a set of punishing regions (defined by the detection range of nodes in the sensor field), and the path that less exposure the robot must be found.

To the best of our knowledge, this is the first work in which both \ac{DOP} and \ac{MEP} formulations are addressed simultaneously as a multi-objective problem, whose application can be extended to a broad range of more realistic contexts.

%%%%%%%%%%%%%%%%%%%%%%%%%%%%%%%%%%%%%%%%%%%%%%%%%%%%%%%%%%%%%%
%%%%%%%%%%%%%%%%%%%%%%%%%%%%%%%%%%%%%%%%%%%%%%%%%%%%%%%%%%%%%%
\section{Preliminaries}
\label{sec:preliminaries}

%%%%%%%%%%%%%%%%%%%%%%%%%%%%%%%%%%%%%%%%%%%%%%%%%%%%%%%%%%%%%%
%\subsection{Point-to-point path}
\subsection{Dubins curve}

%In this paper, we address a minimal exposure orienteering problem focusing on vehicles with nonholonomic constraints.

The Dubins model \cite{Dubins1957Curves} considers vehicles that move in $\Rset^2$ along a path that respects a bounded curvature $\eta$,~i.e.:
\begin{gather*}%\label{eq:dubins}
    \dot{\vehicleconf{}} = 
    \begin{bmatrix}
        \dot{x} \\ \dot{y} \\ \dot{\theta}
    \end{bmatrix}
    = 
    \linvel
    \begin{bmatrix}
        \cos \vehicleori{} \\ \sin \vehicleori{} \\ u/\rho
    \end{bmatrix},
\end{gather*}
\noindent where $\linvel \in \Rset^+$ is a constant linear speed and $u \in \{-1, ~0, ~1\}$. The curvature $\eta$ is inversely proportional to the minimum
turning radius $\rhomin$ the vehicle is capable of executing. %Here, $\rhomin$ represents the vehicle's minimum turning radius constraint.

This model is broadly adopted in robotics since it encompasses a large class of nonholonomic vehicles that range from Ackerman steering cars to fixed-wing aircrafts.

%Such a model encompasses a large class of nonholonomic systems, ranging from Ackermann steering cars to fixed-wing aircrafts. %\ac{UAVs}.

\begin{definition}
A Dubins curve $\dubinscurve(\vehicleconf{i},\vehicleconf{j})$ between two configurations $\vehicleconf{i}$ and $\vehicleconf{j}$ represents the shortest feasible path for a vehicle with minimum turning radius $\rhomin$ \cite{Dubins1957Curves}, and its length is given by $\dubinslength: \SE{2} \times \SE{2} \rightarrow \Rset^+$.
\end{definition}
%
%\aaneto{Creio que isso não precisa ser uma definição.}
%

% On the other hand, given a sequence of configurations $\configurationset = \left\{ \vehicleconf{1}, \ldots, \vehicleconf{\nwaypoints} \right\}$, the shortest Dubins length over $\configurationset$ can be calculated as:
% %
% \begin{equation}
% 	\dubinscircuit = \sum_{i=1}^{\nwaypoints-1} \dubinslength(\vehicleconf{i}, \vehicleconf{i+1}).
% 	\label{eq:dubins_path}
% \end{equation}

% In the \acd{DOP} \cite{Penicka2017DOP}, we must define a subset of $\configurationset$ such that $\dubinscircuit$ remains above a predefined budget $\budget$, maximizing the reward collected in the configurations positions.

%%%%%%%%%%%%%%%%%%%%%%%%%%%%%%%%%%%%%%%%%%%%%%%%%%%%%%%%%%%%%%
%\subsection{Rewarding Targets}

%\aaneto{Explicar brevemente o conjunto de alvos $\positionset$.
%\douglas{Não isso que isso tem que entrar em preliminaries. Não é algo que precisa ser "explicado" antes de prosseguir.}
%}
%\douglas{Aliás, acabei de ver que a seção de Preliminaries sumiu.}

%%%%%%%%%%%%%%%%%%%%%%%%%%%%%%%%%%%%%%%%%%%%%%%%%%%%%%%%%%%%%%
\subsection{Sensing model}

Threatening zones are modeled as a known deployed sensor field $\nodeset = \left\{ \node_1, \node_2, \ldots, \node_\nnodes \right\}$, where $\node_i \in \Rset^2$ is the position of the $i$-th sensor node.
Different sensing models can be found in the literature \cite{Ye2016Hybrid}, from Boolean disks to probabilistic ones. 
We consider the broadly used attenuated disk coverage model since it accurately reflects the detection capability \cite{Nguyen2017Genetic,Binh2019Efficient}.

\begin{definition}%[Sensing function]
    The sensing function $\sfunc(\cdot)$ provides the energy value of a target at a point $\xpos \in \Rset^2$ detected by a single sensor $\node_i$, given by
    \begin{equation}
        \sfunc(\node_i, \xpos) = \frac{\sensingGain}{\|\node_i - \xpos\|^\mu} ,
    \end{equation}
    %\begin{equation*}
    %    \sfunc(\node_i, \xpos) = {\rm e}^{-\alpha \|\node_i - \xpos\|^{\mu}},
    %\end{equation*}
    %
    \noindent where $\sensingGain$ and $\mu$ are known positive constants related to the sensor and environment, while $\|\cdot\|$ is the Euclidean norm.
\end{definition}

% path attenuation exponent depending on the environment and typically ranging from 1.0 to 4.0 [4].

%\douglas{Use detection gain or exposure metrics \cite{Phipatanasuphorn2004Vulnerability}?}

% {
% \color{blue}
% The definition of exposure accounts for the fact that the probability for a target traveling at a constant speed along the path p to be detected by a sensor is proportional to the intensity of the field along p and the length of the path. \cite{Djidjev2007Efficient}
% }

%%%%%%%%%%%%%%%%%%%%%%%%%%%%%%%%%%%%%%%%%%%%%%%%%%%%%%%%%%%%%%
%%%%%%%%%%%%%%%%%%%%%%%%%%%%%%%%%%%%%%%%%%%%%%%%%%%%%%%%%%%%%%
\section{Problem Formulation}
%\subsection{\protect\acd{MEDOP}}
\label{subsec:problem}

%\douglas[red]{Change formulation to open path.}

Let $\positionset = \left\{ \position{1}, \ldots, \position{\nwaypoints} \right\}$ be a set of target locations, $\position{i} \in \Rset^2$, each one with an associated reward $\reward(\position{i}) \in \Rset^+$. The \ac{OP} aims to determine a subset $\positionset_k \subseteq \positionset$ and a sequence of visit that maximizes the accumulated reward, where the visiting path must respect a predefined travel budget $\budget$.
It is assumed that the initial and final positions are fixed and given, and their rewards are $\reward(\position{1}) = \reward(\position{m}) = 0$.

%The \acd{DOP} \cite{Penicka2017DOP} consists of determine a configuration set $\configurationset$ over a subset of $\positionset$ such that the accumulated reward on the total path is maximized.

In the \acd{DOP} \cite{Penicka2017DOP}, the vehicle must assume a particular orientation at each location, and the determination of this configuration is also part of the problem. The final visiting path is a continuous map $\gamma : [0,1] \to \dubinscircuit$, $\gamma(0) = \vehicleconf{1}$ and $\gamma(1) = \vehicleconf{\nwaypoints}$, formed by Dubins curves:
\begin{equation}
	\dubinscircuit = \bigcup_{i=1}^{k-1} \dubinscurve(\vehicleconf{i}, \vehicleconf{i+1}),
	\label{eq:dubins_path}
\end{equation}
where $k$ defines subset $\positionset_k$ of locations that will be visited. The orientation of two Dubins curves that meet at the same position must be equal, and $|\dubinscircuit| = \sum_{i=1}^{k-1} \dubinslength(\vehicleconf{i}, \vehicleconf{i+1})$ denotes the length of the path.

In this paper, the orienteering problem is addressed in the presence of a known deployed sensor field $\nodeset$. %Therefore, before formally defining our main problem, it is necessary to establish some terms of the \ac{MEP} problem.
The concept of \emph{sensor field intensity} is associated with the likelihood of a given target being detected at a location $\xpos$ by any of the sensors nodes \cite{Meguerdichian2001Exposure}. Similar to the sensing function $\sfunc(\cdot)$, there are different forms of estimating the field intensity $\ifunc$, and we consider the \emph{all-sensor field intensity function}, given by:
\begin{equation}
    \ifunc(\nodeset, \xpos) = \sum_{i \in \nodeset} \sfunc(\node_i, \xpos).
\end{equation}

The field intensity function is used to calculate \emph{exposure} $\efunc$, which is related to the probability of a target being detected traveling at constant speed along an arbitrary path \cite{Djidjev2007Efficient}. %It is proportional to the intensity of the field and the length of the path \cite{Djidjev2007Efficient}. 
Given the Dubins path $\dubinscircuit$, it can be written as:
\begin{equation}
    \efunc(\nodeset, \dubinscircuit) = \int_{0}^{|\dubinscircuit|} \ifunc(\nodeset, \dubinscircuit(s)) ds.
    \label{eq:exposure}
\end{equation}

% \aaneto{

% Exposure of a single Dubins curve between $\vehicleconf{a}$ and $\vehicleconf{b}$:

% \begin{equation}
%     \efunc(\nodeset, \dubinscurve(\vehicleconf{a},\vehicleconf{b})) = \int_{0}^{\dubinslength(\vehicleconf{a}, \vehicleconf{b})} \ifunc\Big( \nodeset, \dubinscurve(\vehicleconf{a},\vehicleconf{b}) \Big) ds.
%     %\label{eq:exposure}
% \end{equation}

% Exposure of all Dubins curves:

% \begin{equation}
%     \efunc(\nodeset, \dubinscurve^*) = \sum_{i = 1}^{|\configurationset|-1} \efunc\Big( \nodeset, \dubinscurve(\vehicleconf{i},\vehicleconf{i+1}) \Big)
% \end{equation}
% }

\begin{problem}[Minimal Exposure Dubins \ac{OP}] 
    Given a set of target locations $\positionset=\{\position{i}\}_{i=1}^\nwaypoints$, each location $i$ with a known associated reward $\reward(\position{i})$, and a sensor field formed by a set of nodes $\nodeset=\{\node_i\}_{i=1}^\nnodes$. Determine a path $\dubinscircuit$ that visits a subset of locations $\positionset_k \subseteq \positionset$ to maximize the total collected reward, whilst minimizes the exposure. Formally:
    \begin{eqnarray}
    	\max_{~\dubinscircuit} & \displaystyle \sum_{\position{} \in \positionset_k} \reward(\position{}),
    	\\ [10pt]
    	\min_{~\dubinscircuit} & \efunc(\nodeset, \dubinscircuit),
    	\\ [10pt]
    	\textrm{subject to} & ~|\dubinscircuit| \leq \budget.
    \end{eqnarray}
    \label{prob:MEDOP}
\end{problem}

% \aaneto{$\min_{\nodeset, \configurationset}$ ~~ $\efunc(\nodeset, \dubinscurve^*)$}

% \daigo{Given a set of locations $\positionset=\{\position{i}\}_{i=1}^m$ with associated reward $\reward(\position{})$ and a sensor field formed by a set of nodes $\nodeset$, find a path $\path$ that visits a set of locations $\positionset_\text{visit}(\path)\subseteq \positionset$ to maximize the collected reward while minimizing the exposure: 
% \begin{eqnarray}
% \max_{\path} \sum_{\position \in \positionset_\text{visit}(\path)} \reward(\position{})\\
% \min_{\boldsymbol{\gamma}}  E(\nodeset,\boldsymbol{\gamma})\\
% \text{s.t.\;\;} |\boldsymbol{\gamma}| < T_\text{budget},
% \end{eqnarray}
% where $|\boldsymbol{\gamma}|$ denotes the length of the path.
% }
%\douglas{Leave as a multi-objective problem or simplify?}
%\douglas{Consider a weighted linear combination?}

This problem (simplified as to define $\dubinscircuit$) involves the selection of the $k$ locations that will compose $\positionset_k$ and their permutation, as well as the appropriate orientation at each location. Furthermore, we have to design the path in such a way that it avoids the threatening zones, in our case, by locally adjusting the turning radius.

%%%%%%%%%%%%%%%%%%%%%%%%%%%%%%%%%%%%%%%%%%%%%%%%%%%%%%%%%%%%%%
%%%%%%%%%%%%%%%%%%%%%%%%%%%%%%%%%%%%%%%%%%%%%%%%%%%%%%%%%%%%%%
\section{Methodology}
\label{sec:methodology}

The \acd{MEDOP} is constituted by both combinatorial and continuous optimization components. The combinatorial part of the problem is related to the subset selection of the most rewarding locations and the determination of the sequence of visits. The continuous part deals with the assignment of the vehicle's orientation at the locations and the selection of a feasible and convenient turning radius, where both greatly affect the path shape.

%The \acd{MEDOP} is constituted of both combinatorial and continuous optimization components. The combinatorial portion is related to the selection of the most rewarding locations and the determination of the visiting sequence, while the continuous part deals with the assignment of the orientation at the locations and with the selection of a convenient and feasible turning radius, which determines the path shape.

Considering the high dimensionality of the search space, the problem is quite intractable, and finding efficient methods is an open challenge. In this context, evolutionary-based approaches proved to be a suitable option. \ac{GA} mimics evolutionary processes found in nature, working to find the fittest individual after a series of operations such as selection, reproduction, and mutation.

In multi-objective optimization problems, the usual concept of a single optimum solution (fittest individual) does not apply directly. Usually, the goal is to obtain different solutions with as many associated values as objectives. The best solutions for such problems belong to what is called a Pareto front, where all options are considered acceptable and equally good.

An important difference between multi-objective evolutionary algorithms over single-objective ones is in how to compare individuals, necessary in the selection phase. The selection is based on the \emph{dominance} between individuals, given the objective dimensions. Usually, it is used a tournament selection considering a distance operator combined with a Pareto-based sorting (e.g., NSGA-II,~SPEA2,~etc.).

%, as proposed by the NSGA-II \cite{Deb2002Fast}.

%s\aaneto{Comentar algo sobre o porque dessa escolha.}

%%%%%%%%%%%%%%%%%%%%%%%%%%%%%%%%%%%%%%%%%%%%%%%%%%%%%%%%%%%%%%
\subsection{Encoding}

% The chromosome is encoded as a list of tuples
% %
% \begin{equation}
%     \sequence{\position{\permutation_1},\vehicleori{1},\rho_1}, \sequence{\position{\permutation_2},\vehicleori{2},\rho_2}, \ldots, \sequence{\position{\permutation_\nwaypoints},\vehicleori{\nwaypoints}, \rho_\nwaypoints}
%     \label{eq:chromosome}
% \end{equation}
% %
% \noindent representing the permutation of the points to be visited $\position{\permutation_i}$, as well as the corresponding orientation $\vehicleori{i}$ and curvature radius $\rho_i$ used to compute the connecting Dubins' curves. The indices of the genes correspond to the positions that must be visited.

% The visiting sequence is determined by using a random-key scheme \cite{Bean1994Genetic}, where each gene (tuple) in the chromosome has a parameter $\permutation \in (0,1)$ defining its order, uniformly selected during the initialization. A negative value will be assigned to locations that are not part of the solution, i.e., they will not be visited.
% %
% Also, the vehicle orientation at a given location $\vehicleori{} \in [0, 2\pi)$ and the turning radius used to determine the curve to the next position of the sequence $\rho \in [\rhomin, \rhomax]$.

The chromosome is encoded as an array of tuples (Fig. \ref{fig:chromosome}), which represents a permutation of the locations to be visited as well as specific parameters to the calculation of the connecting Dubins curves. The indices of the genes correspond to the positions that must be visited.

\begin{figure}[h]
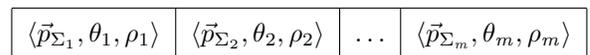


  \renewcommand{\tabcolsep}{2mm} 
  \renewcommand{\arraystretch}{1.5}
  \centering
  \vspace{.15cm}
  %$\chromosome{}$ \quad = \quad
  \begin{tabular}{|c|c|c|c|c|c|c|}
  \hline
   $\sequence{\position{\permutation_1},\vehicleori{1},\rho_1}$ & $\sequence{\position{\permutation_2},\vehicleori{2},\rho_2}$ & $\ldots$ & $\sequence{\position{\permutation_\nwaypoints},\vehicleori{\nwaypoints}, \rho_\nwaypoints}$ \\
  \hline
  \end{tabular}
  \vspace{.15cm}
  \caption{Chromosome structure.}
  \label{fig:chromosome}
  
\end{figure}

The definition of $\positionset_k$ and the permutation are based on a random-key scheme \cite{Bean1994Genetic}. Each gene (tuple) in the chromosome has a parameter $\permutation \in [0,1]$ that defines the order, and which is uniformly selected at random during initialization. A negative value will be used to represent that the location is not part of the solution, i.e., will not be visited. The start and goal locations receive $\permutation_1 = 0$ and $\permutation_\nwaypoints=1$.

The gene also has $\vehicleori{} \in [0,2\pi)$ and $\rho \in [\rhomin, \rhomax]$, which represent the orientation the vehicle must assume at a given location and the value of the turning radius used to determine the curve to the next location in the sequence, respectively.

%%%%%%%%%%%%%%%%%%%%%%%%%%%%%%%%%%%%%%%%%%%%%%%%%%%%%%%%%%%%%%
\subsection{Fitness evaluation}

% At first, the chromosome must be decoded to obtain a solution. First, only genes with $\permutation > 0$ have been considered, and they are sorted in ascending order according to their random keys. The remaining genes are assumed to represent a valid solution (respecting $\budget$ constraint), since this validation is done during the generation of the offspring.
% %
% Next, a path is constructed accordingly to Eq.~\eqref{eq:dubins_path}, with each segment $\dubinscurve(\vehicleconf{i},\vehicleconf{i+1})$ being determined by the gene-specific values for the orientation and minimum turning radius.
% %
% Finally, the fitness of the individual is composed of the path length $\dubinscircuit$, and the exposure value  $\efunc(\cdot)$ given by Eq.~\eqref{eq:exposure}.
% \aaneto{o fitness tem dois valores?}

Initially, we must decode the chromosome to obtain a solution (visit path). Hereupon, genes with $\permutation = -1$ are discarded. We assume the remaining genes represent a valid solution, i.e., a path that respects $\budget$, as this validation is done during the generation of the offspring.

First, the genes are sorted by their random keys in ascending order.
Next, we construct the path accordingly to Eq.~\eqref{eq:dubins_path}. Each segment will be determined by considering the gene-specific values for the orientation and turning radius. 

Finally, the fitness of the individual comprises the collected reward and the path's exposure, determined by Eq.~\eqref{eq:exposure}.

%%%%%%%%%%%%%%%%%%%%%%%%%%%%%%%%%%%%%%%%%%%%%%%%%%%%%%%%%%%%%%
\subsection{Crossover}

% This step is responsible for the exploitation of the already known solutions, repeating decisions (in our case, exchanging individual genes) that have worked well so far.
% %
% It will only exchange the values of the attributes, but not the index of the location they were assigned to. It allows us to apply different types of operators and not only order-restricted ones. The well-known two-point crossover is a suitable and convenient option in our case, as it can produce changes considering portions of the individual. % (Fig. \ref{fig:crossover}).
% %
% % \begin{figure}[htpb]
% %     \centering
% %     \includegraphics[width=.8\linewidth]{img/crossover.pdf}
% %     \caption{Illustration of the two-point crossover operator.}
% %     \label{fig:crossover}
% % \end{figure}
% %
% Then, we evaluate the validity of each new individual and adjust it if necessary. If the travel budget limit is violated, we repeatedly remove at random nodes that belong to the path, i.e., assign $\permutation = -1$, until feasibility is achieved.

The crossover is responsible for the exploitation of the already known solutions, i.e., repeat decisions (in our case exchange individual genes) that have worked well so far.

This step will only exchange the values of the attributes, but not the index of the location a gene assigned to. This allows us to apply different types of operators and not only order-restricted ones. The well-known two-point crossover is a suitable and convenient option in our case, as it can produce changes considering only portions of the individual. % (Fig. \ref{fig:crossover}).

% \begin{figure}[htpb]
%     \centering
%     \includegraphics[width=.8\linewidth]{img/crossover.pdf}
%     \caption{Illustration of the two-point crossover operator.}
%     \label{fig:crossover}
% \end{figure}

Then, we evaluate the validity of each new individual and adjust it if necessary. If the travel budget limit is violated, we repeatedly remove at random nodes that belong to the path, i.e., assign $\permutation = -1$, until feasibility is achieved.

%%%%%%%%%%%%%%%%%%%%%%%%%%%%%%%%%%%%%%%%%%%%%%%%%%%%%%%%%%%%%%
\subsection{Mutation}

% This step brings the innovation (exploration of the search space) and occurs internally to the individual.
% %
% In our context, it is responsible to randomly determine new values to the order of visit, to the orientation at a certain location, and to the turning radius of the curve connecting two consecutive points in the sequence.

% A different permutation is achieved by assigning a new value for $\permutation$ in the $(0,1)$ interval. The start and goal locations do not suffer any change, and their values remain constant, i.e., $\permutation_0 = 0$ and $\permutation_\nwaypoints=1$, throughout the generations.

The mutation step brings innovation (exploration of the search space) and occurs internally to the individual.

In our context, it is responsible to randomly select new values to the order of visit, the orientation at a certain location, and the turning radius of the curve connecting two consecutive points in the sequence.

A different permutation is achieved by assigning a new value for $\permutation$ in the $(0,1)$ interval. The start and goal locations do not suffer any change, and their values remain the same as they were initialized throughout the generations.

%first two are attributes commonly manipulated in DTSP-related techniques, and will be updated accordingly to $\permutation \sim \mathcal{U}(0, 1)$ and $\vehicleori{} \sim \mathcal{U}(0, 2\pi)$.

The heading $\vehicleori{}$ will be updated accordingly to a von Mises (circular normal) distribution with mean $\mu$ and dispersion $\kappa$:
\begin{equation}
    p(x) = \frac{e^{\kappa \cos(x-\mu)}}{2\pi I_0(\kappa)},
\end{equation}
where $I_0(\kappa)$ is the modified Bessel function of order 0.

In our case, the mean $\mu$ will be defined as the current orientation. This distribution was chosen as it provides a good balance between the probability of searching in its vicinity and attempting a completely different new value.

Another innovative aspect of our approach is the variation of the turning radius along the path. New values will be uniformly chosen at random in an interval defined by the vehicle-specific constraint ($\rhomin$) and a scenario-related one ($\rhomax$). This is key to minimize the exposure of a path, however, the increase in the length can impact the collected reward (less visited locations) given the travel budget.

Finally, after the new values are assigned, the same budget violation check and correction operation mentioned in the previous step are executed to avoid invalid individuals.

%%%%%%%%%%%%%%%%%%%%%%%%%%%%%%%%%%%%%%%%%%%%%%%%%%%%%%%%%%%%%%
%%%%%%%%%%%%%%%%%%%%%%%%%%%%%%%%%%%%%%%%%%%%%%%%%%%%%%%%%%%%%%
\section{Experiments}
\label{sec:experiments}

%In this section, we evaluate and discuss different aspects of the methodology.

The simulation framework was implemented with Python~3.7 and uses the DEAP \cite{DEAP_JMLR2012} library.
The sensing function for all sensors considers coefficients $\sensingGain=50$ and $\mu=2$. We also apply a threshold where $\sfunc(\cdot) > 30$.

% {
% \color{blue}
% The coefficients Ki and $\alpha_i$ were chosen as 10 and 2 respectively for all sensors \cite{Ferrari2010Potential}.
% }

%%%%%%%%%%%%%%%%%%%%%%%%%%%%%%%%%%%%%%%%%%%%%%%%%%%%%%%%%%%%%%
\subsection{Dataset}
\label{sec:dataset}

Since this is the first time the problem is being tackled, we propose two new test instances to the \ac{MEDOP} and make them available\footnote{\url{https://www.dcc.ufmg.br/~doug/medop/}}.
\begin{itemize}
    \item \textbf{Scenario A:} A 30m x 22m environment, with 11 nodes displaced in a \emph{cross} shape. There are 18 locations to be visited, with rewards in $\{0.2, 0.4, 0.6, 0.8, 1.0\}$.
    \item \textbf{Scenario B:} A 30m x 30m environment, with 8 nodes displaced in a \emph{grid} formation. There are 15 locations to be visited, with rewards in $\{0.2, 0.4, 0.6, 0.8, 1.0\}$.
\end{itemize}

\subsection{Illustrative example}

Initially, we present an illustrative example for better visualization of the test instances and understanding of the achieved results\footnote{Video of the execution: \url{https://youtu.be/5lyYZVc5weU}}.
In this example, we consider $\budget = 100$ and $\rho \in [1.0, 2.0]$ for both scenarios. Table \ref{tab:parameters} summarizes the parameters used for our evolutionary algorithm.

\begin{table}[htpb]
  \centering
  \caption{Algorithm parameters.}
  \label{tab:parameters}
  \begin{tabular}{ll}
  \toprule
  \textbf{Parameter}    & \textbf{Value}       \\
  \midrule
  Population size       & 400                  \\
  Number of generations & 400                  \\
  Selection method      & NSGA-III (reference-point)\\
  Crossover probability & 0.8                 \\
  Crossover operator    & Two-point crossover    \\
  Mutation probability (individual) & 0.4 \\
  Mutation probability (gene)  & 0.02 \\
  Dispersion (von Mises)           & $\kappa = 2.0$ \\
  \bottomrule
  \end{tabular}
\end{table}

Figures \ref{fig:medop-a} and \ref{fig:medop-b} show different solutions for  the \textbf{Scenario~A} and \textbf{Scenario B} instances, respectively. These solutions are in different parts of the final Pareto front, and it is possible to observe the trade-off between Reward and Exposure. Table~\ref{tab:minmax} presents the extreme values of the frontier for each instance.

\begin{figure*}[htpb]
    \centering
    \subfloat[]{\includegraphics[height=4cm]{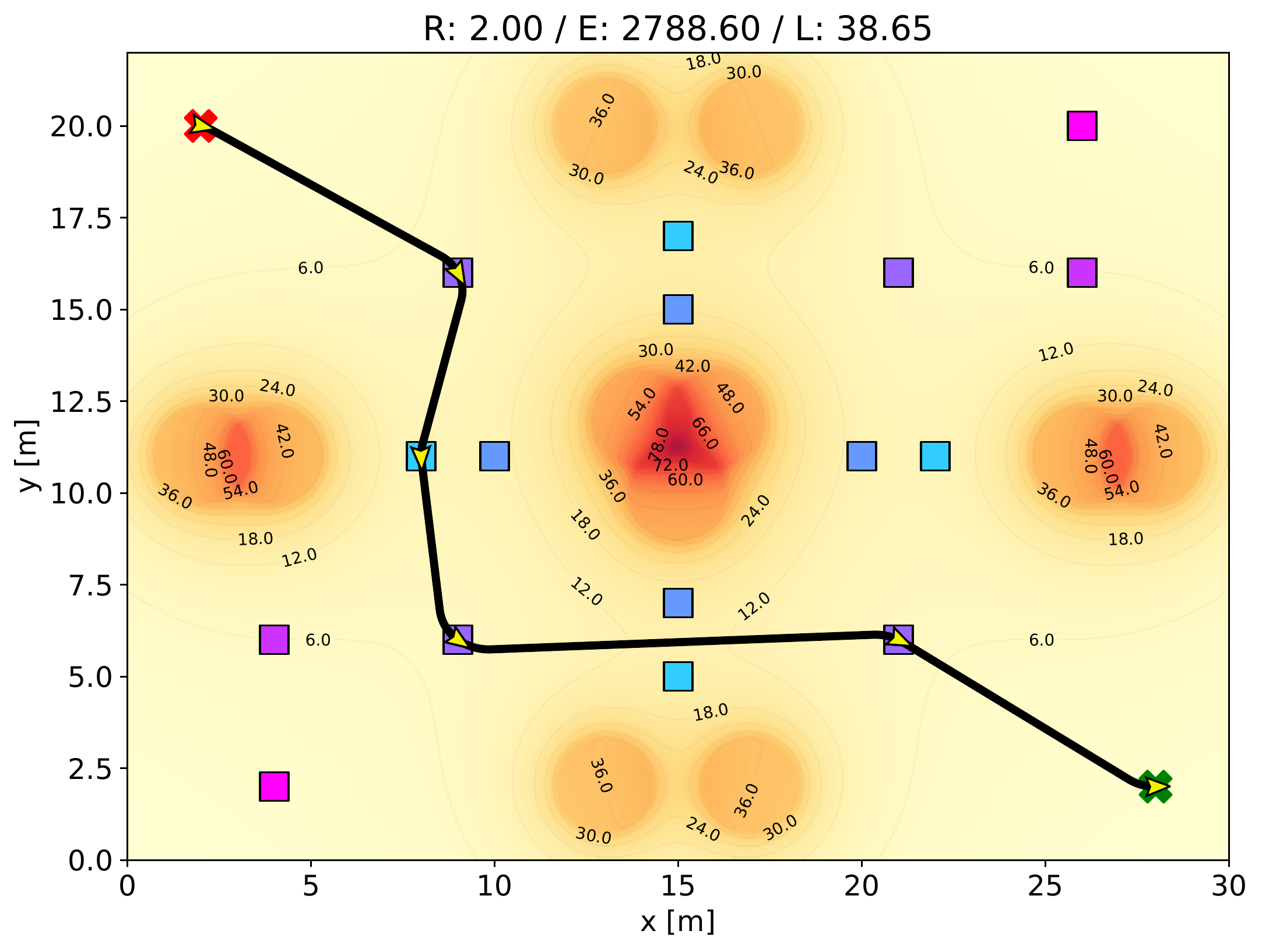}
    }
    \subfloat[]{{\includegraphics[height=4cm]{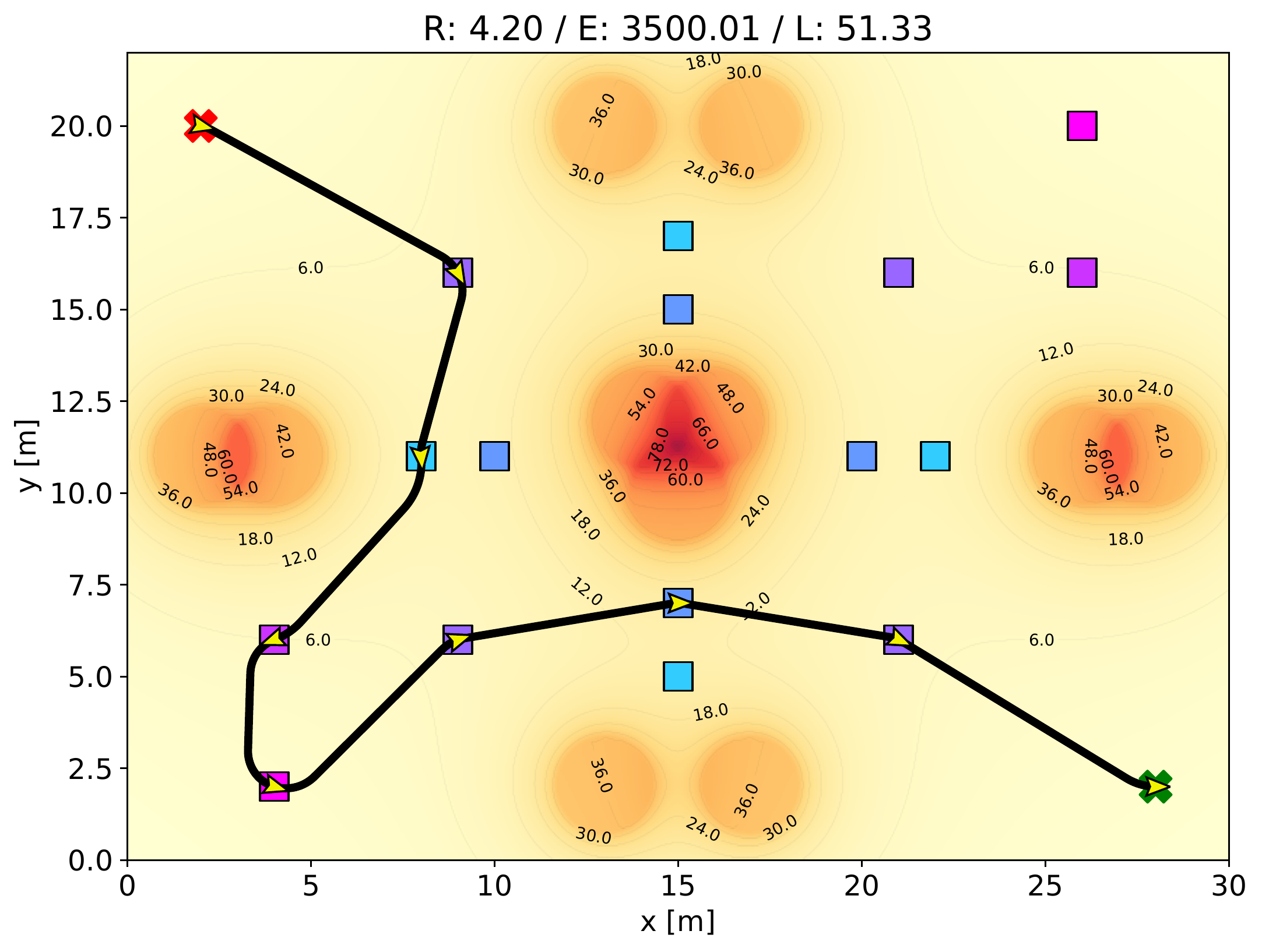}}
    }
    \subfloat[]{{\includegraphics[height=4cm]{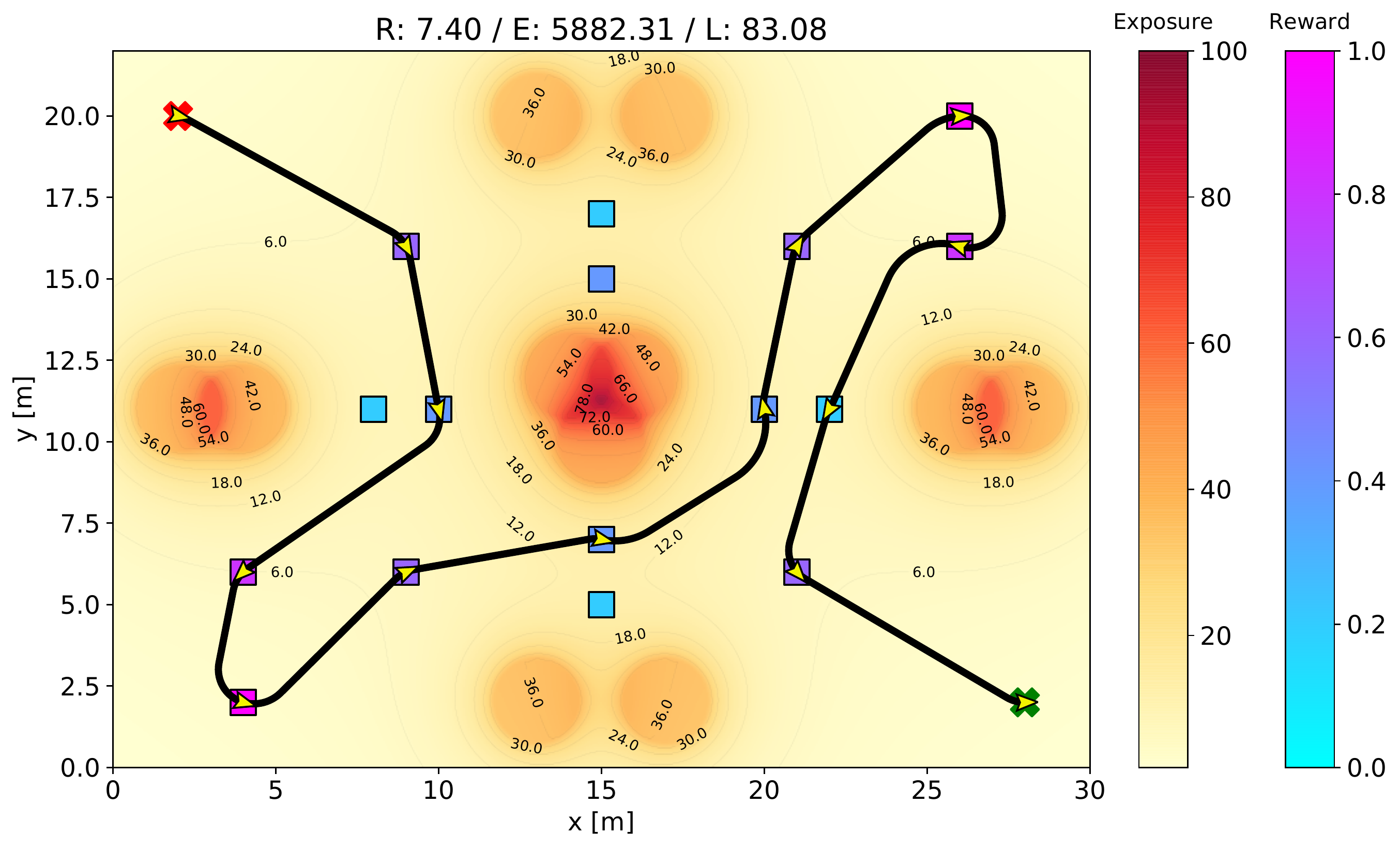}}
    }    
    \caption{Example solutions along the Pareto front for Scenario A ($\budget = 100$, $\rho \in [1.0, 2.0]$). The title shows the values for Reward (R), Exposure (E), and Length (L) of the solutions.}
    \label{fig:medop-a}
    \vspace{-2mm}
\end{figure*}

\begin{figure*}[htpb]
    \centering
    \subfloat[]{\includegraphics[height=4cm]{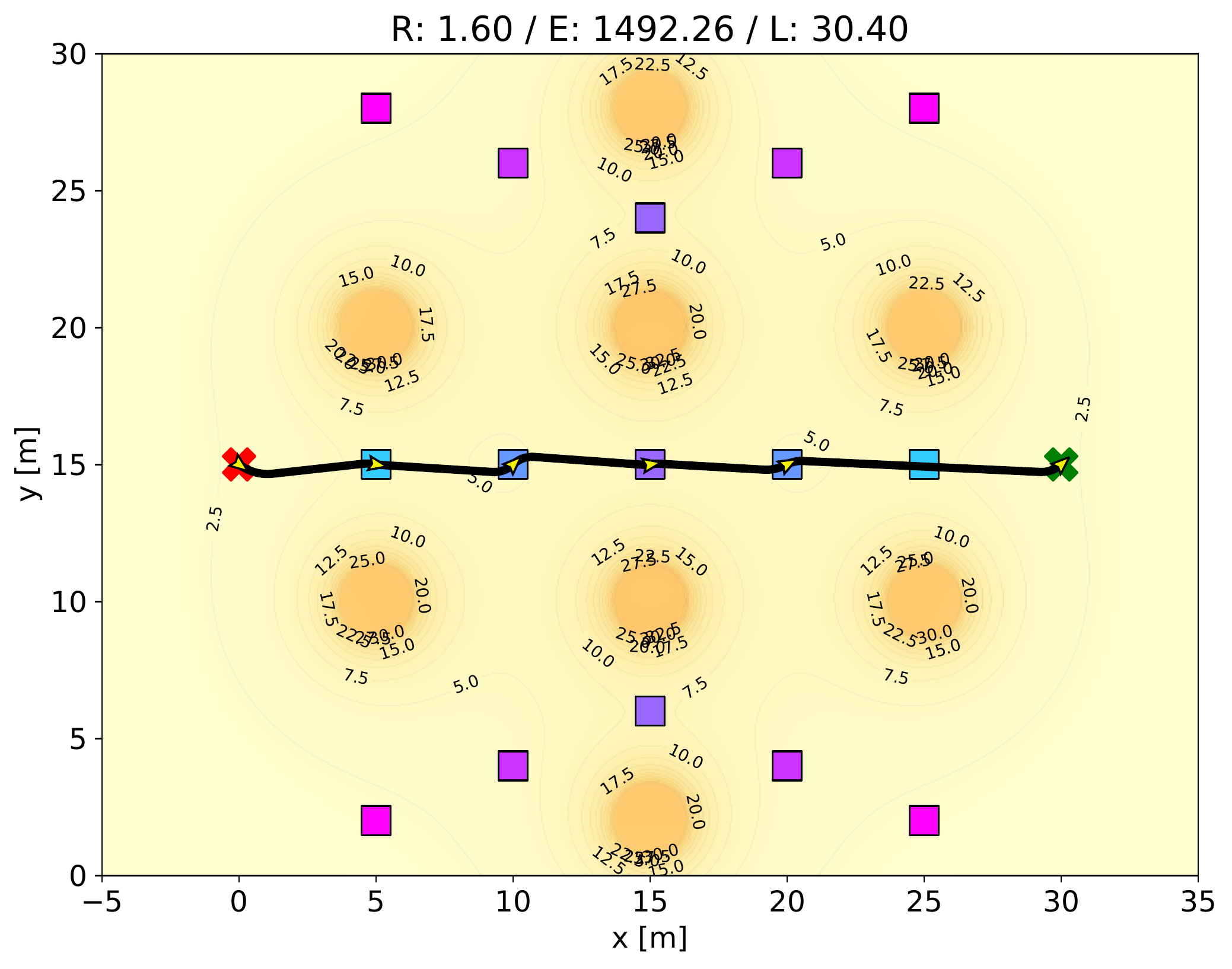}
    }
    \hspace{1mm}
    \subfloat[]{{\includegraphics[height=4cm]{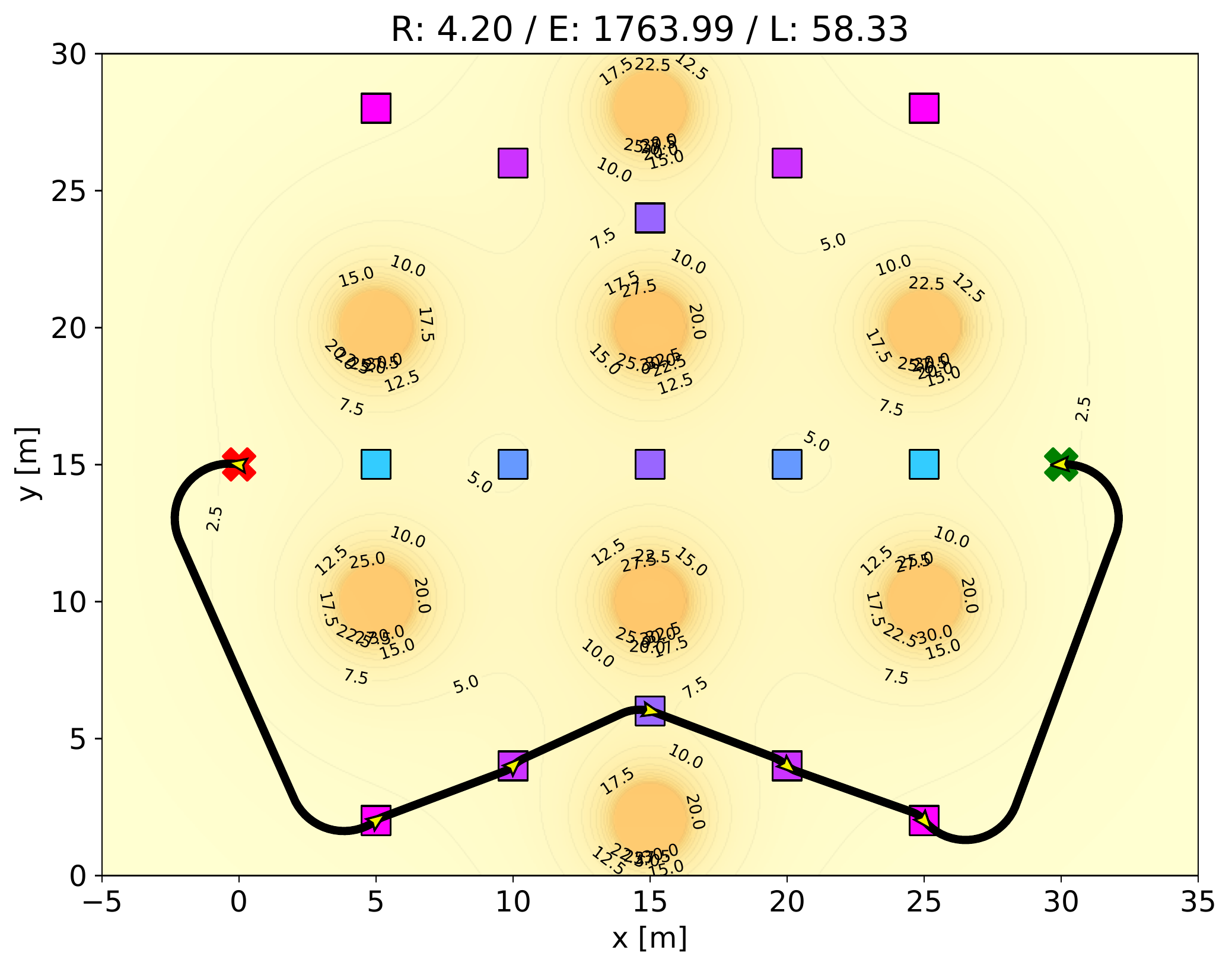}}
    }
    \hspace{1mm}
    \subfloat[]{{\includegraphics[height=4cm]{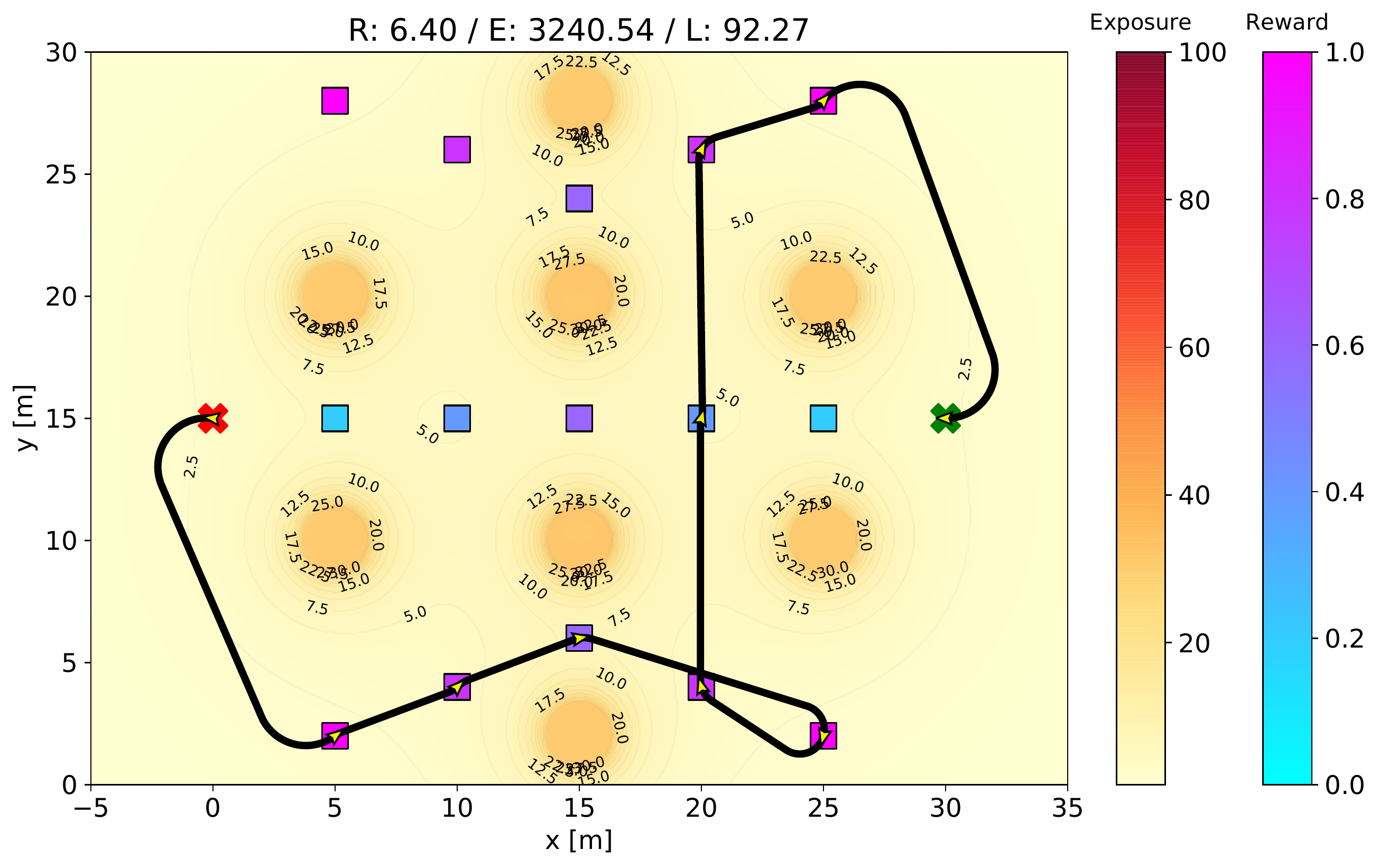}}
    }    
    \caption{Example solutions along the Pareto front for Scenario B ($\budget = 100$, $\rho \in [1.0, 2.0]$). The title shows the values for Reward (R), Exposure (E), and Length (L) of the solutions.}
    \label{fig:medop-b}
    \vspace{-4mm}
\end{figure*}

\begin{table}[htpb]
  \centering
  \caption{Pareto front extreme solutions.}
  \label{tab:minmax}
  \begin{tabular}{@{\extracolsep{5pt}}lcccc}
  \toprule
      & \multicolumn{2}{c}{\textbf{Scenario A}} & \multicolumn{2}{c}{\textbf{Scenario B}}       \\
      \cline{2-3} \cline{4-5}
      & min & max & min & max \\
  \midrule
  \textbf{Reward}       & 1.40 & 7.60 & 0.80 & 7.20 \\
  \textbf{Exposure}     & 2682.81 & 6671.75 & 1491.30 & 5440.37  \\ 
  \textbf{Length}     & 36.67 & 88.91 & 30.18 & 99.60  \\ 
  \bottomrule
  \end{tabular}
\end{table}

% \begin{figure}[htpb]
%     \centering
%     \subfloat[MEDOP A]{\includegraphics[width=.9\linewidth]{img/MEDOP-A-v2-Pareto.pdf}
%     }
%     \\
%     \subfloat[MEDOP B]{{\includegraphics[width=.9\linewidth]{img/MEDOP-B-v8-Pareto.pdf}}
%     }
%     \caption{ToDo.}
%     \label{fig:example-pareto}
% \end{figure}

When tackling with multi-objective optimization problems, any of the Pareto front solutions are similarly good and acceptable. Therefore, a decision-making process can use different criteria to select the most adequate one, and the goal of a  proper methodology for such problems is to provide it with a reasonable amount of rather distinct options.

For comparison purposes, we also solve the same instances considering only the reward objective ($\nodeset = \emptyset$). As can be seen in Fig. \ref{fig:medop-single}, by not taking into account the exposure restriction we can achieve a higher collected reward. Furthermore, the best individuals tend to keep $\rho$ close too $\rhomin$, as this helps to add more locations to the path.

\begin{figure}[htpb]
    \centering
    \subfloat[MEDOP A]{\includegraphics[width=.48\linewidth]{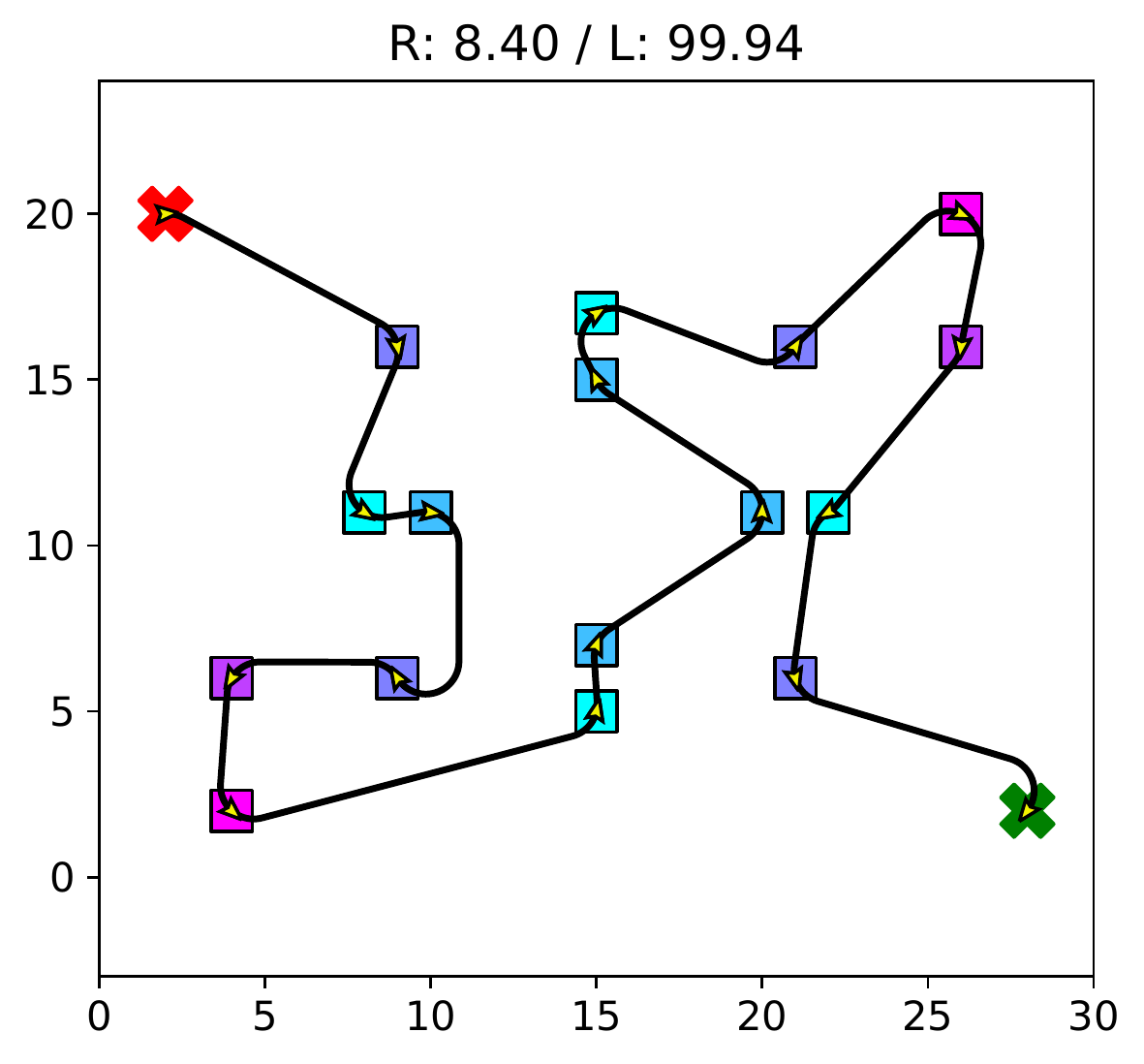}
    }
    \subfloat[MEDOP B]{{\includegraphics[width=.48\linewidth]{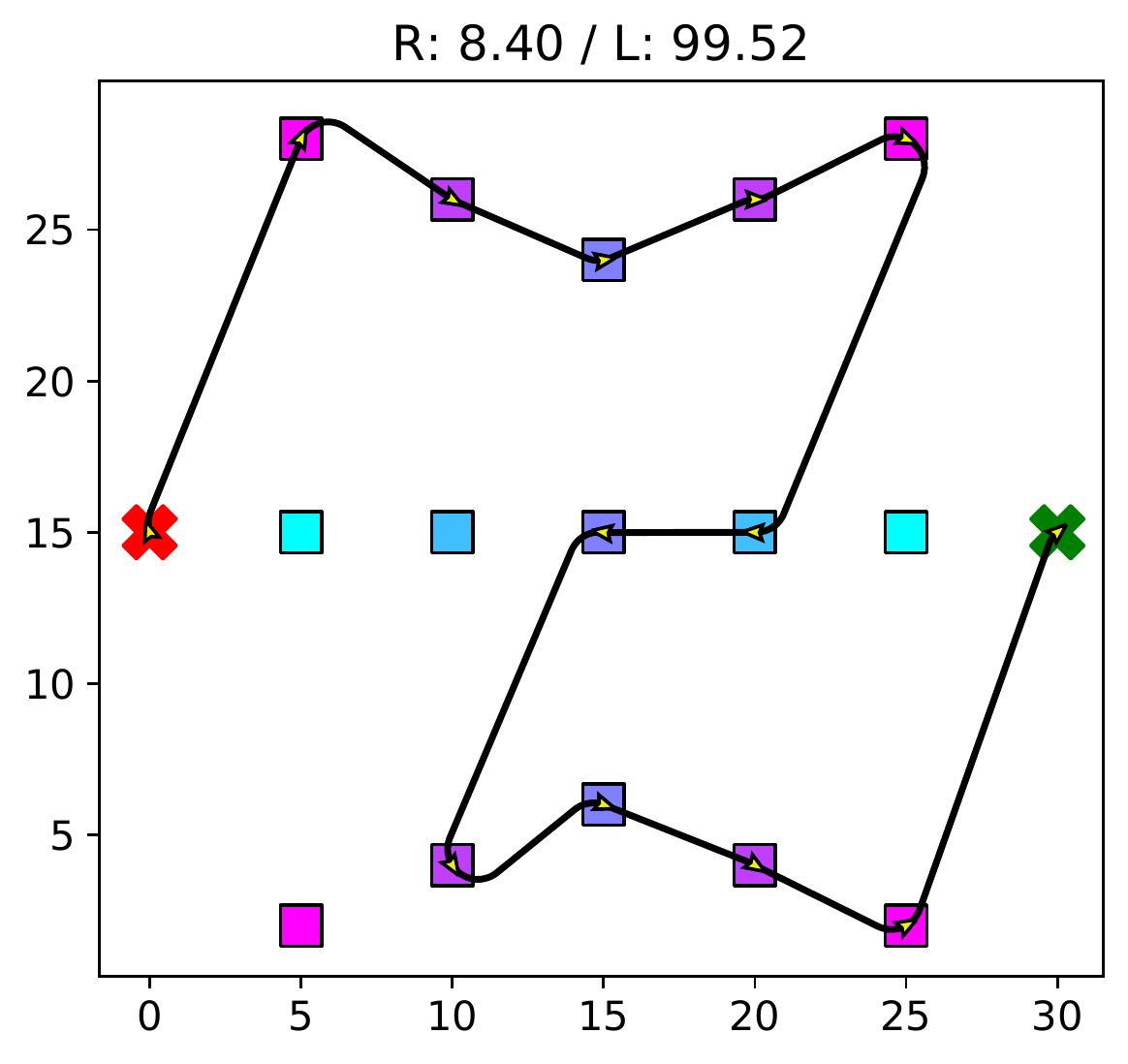}}
    }
    \caption{Single-objective solutions ($\budget = 100$, $\rho \in [1.0, 2.0]$).}
    \label{fig:medop-single}
\end{figure}

A discussion about the use of the methodology for the single-objective case is presented in the next section.

\subsection{Single-objective OP}
\label{sec:singleexp}

Although our method is also applicable to the single-objective OP case, it was not designed for that purpose. %The proposed genetic operators might be too simplistic for more dense and complex scenarios.
Since we are tackling with multiple objectives, the final result is a set of optimal solutions. Hence, the goal is not only the convergence to an optimal solution but also the diversity of the solutions, which might assist a decision-making step.

Figure \ref{fig:example_heuristic} presents a result of our method for the well-known Set~66~\cite{Chao1996Fast} dataset, considering $\budget = 130$ and a fixed value of $\rho = 0.7$. Figure \ref{fig:example_heuristic_a} shows the result using the methodology as proposed (with $\kappa = 8.0$). The genetic operators might be too simplistic for more dense and complex scenarios (orientations are challenging). Figure \ref{fig:example_heuristic_b} shows the result considering a new operation during the mutation that simply tries to align the orientations of consecutive locations. 

\begin{figure}[htpb]
    \centering
    \subfloat[Methodology]{\includegraphics[width=.48\linewidth]{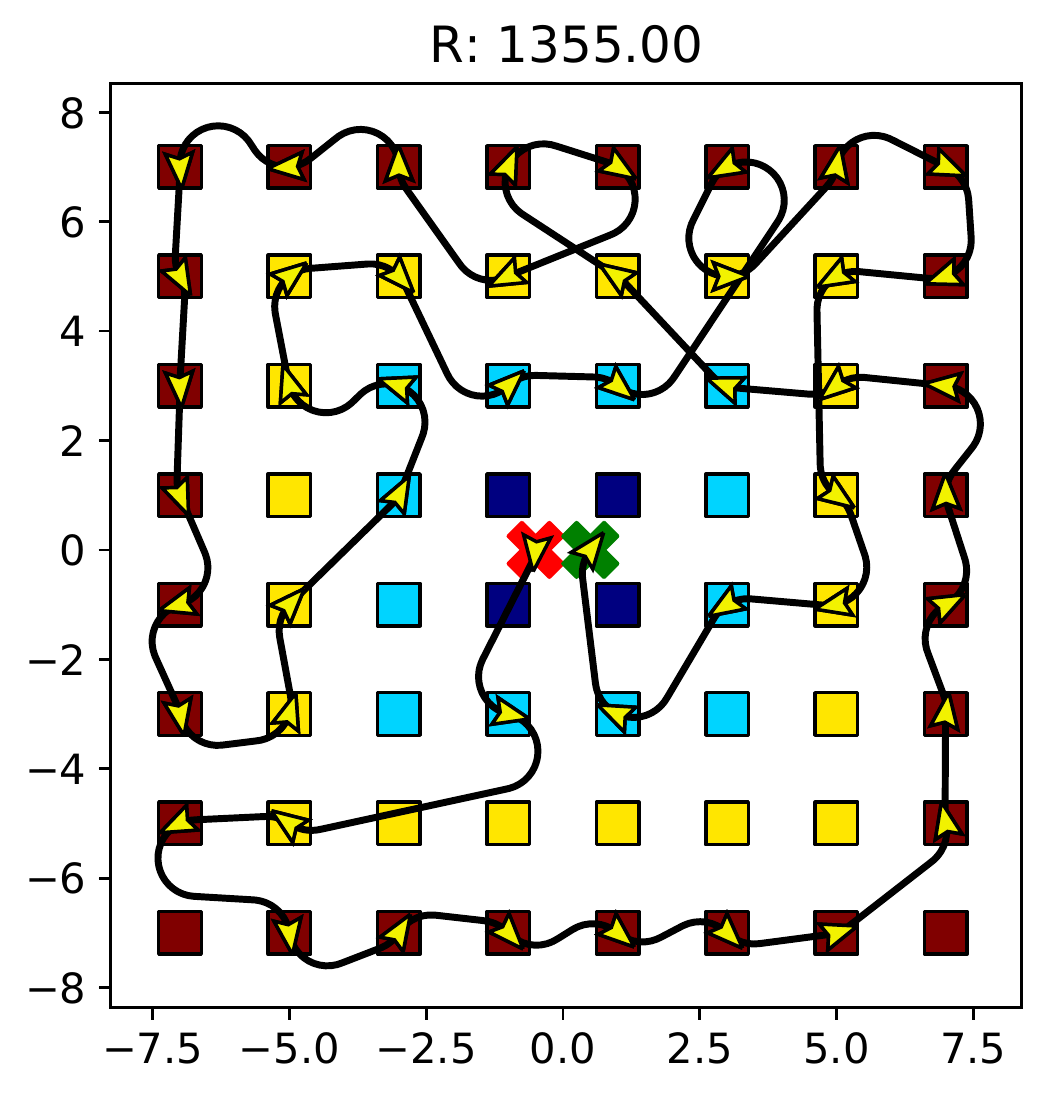}
    \label{fig:example_heuristic_a}
    }
    \subfloat[Methodology+Alignment]{{\includegraphics[width=.48\linewidth]{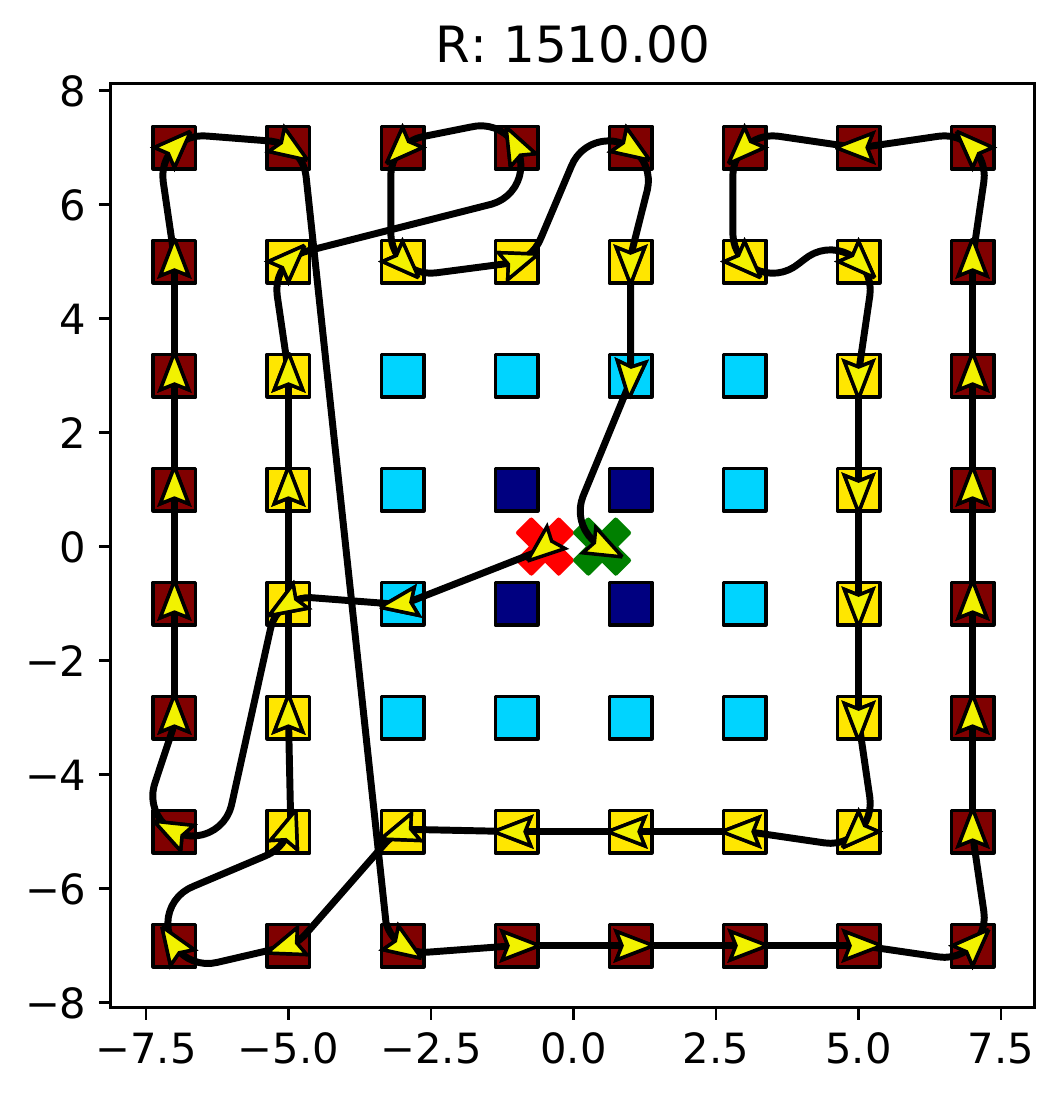}}
    \label{fig:example_heuristic_b}
    }
    \caption{Results for Set~66~\cite{Chao1996Fast} dataset ($\budget = 130$, $\rho = 0.7$).}
    \label{fig:example_heuristic}
\end{figure}

Table \ref{tab:set66} compares our results with other approaches in the literature. As can be seen, we achieve about $80\%-90\%$ of the state-of-the-art, however, with a simpler approach and without considering the assist of local search procedures.

\begin{table}[htpb]
\centering
\caption{Results for Set 66 dataset ($\budget = 130$, $\rho = 0.7$).}
\label{tab:set66}
\begin{tabular}{cccc}
\toprule
\textbf{VNS} \cite{Penicka2017DOP} & \textbf{Hybrid GA} \cite{Tsiogkas2018DCOP} & \textbf{Ours} & \textbf{Ours} (aligned) \\
\midrule
 1675 & 1670 & 1355 & 1510 \\
 \bottomrule
\end{tabular}
\end{table}

Besides the satisfactory results when compared to other methods, in our case it is not recommended to explicitly benefit one objective over the other, for example, by applying the alignment strategy \cite{Macharet2014Orientation}. However, this shows the method is adaptable to easily incorporate more focused operators.

\subsection{Numerical analysis}

In this section, we investigate the effectiveness of our approach accordingly to a significant number of simulations. The behavior will be evaluated considering different combinations for the travel budget and turning radius interval.

The experiments were performed considering both test instances presented in Sec. \ref{sec:dataset}. We vary the parameters accordingly to the following: $\budget = \{60, 80, 100, 120\}$ and $\rhomax = \{1.0, 2.0, 3.0, 4.0\}$, with a fixed $\rhomin = 1.0$.
For each case, the algorithm is executed 30 times, and we select the solution with the highest reward and smallest exposure. The overall performance is assessed through the final Pareto~front. 

Figure \ref{fig:multi-pareto-fixed-rhomax} shows the different results for a fixed $\rhomax = 2.0$ and different values for $\budget$. As expected, a greater budget allows for the collection of more rewards. The behavior of all scenarios is similar, as there is not much room for different solutions with the fixed and equal turning radius interval.

\begin{figure}[htpb]
    \centering
    \includegraphics[width=.8\linewidth]{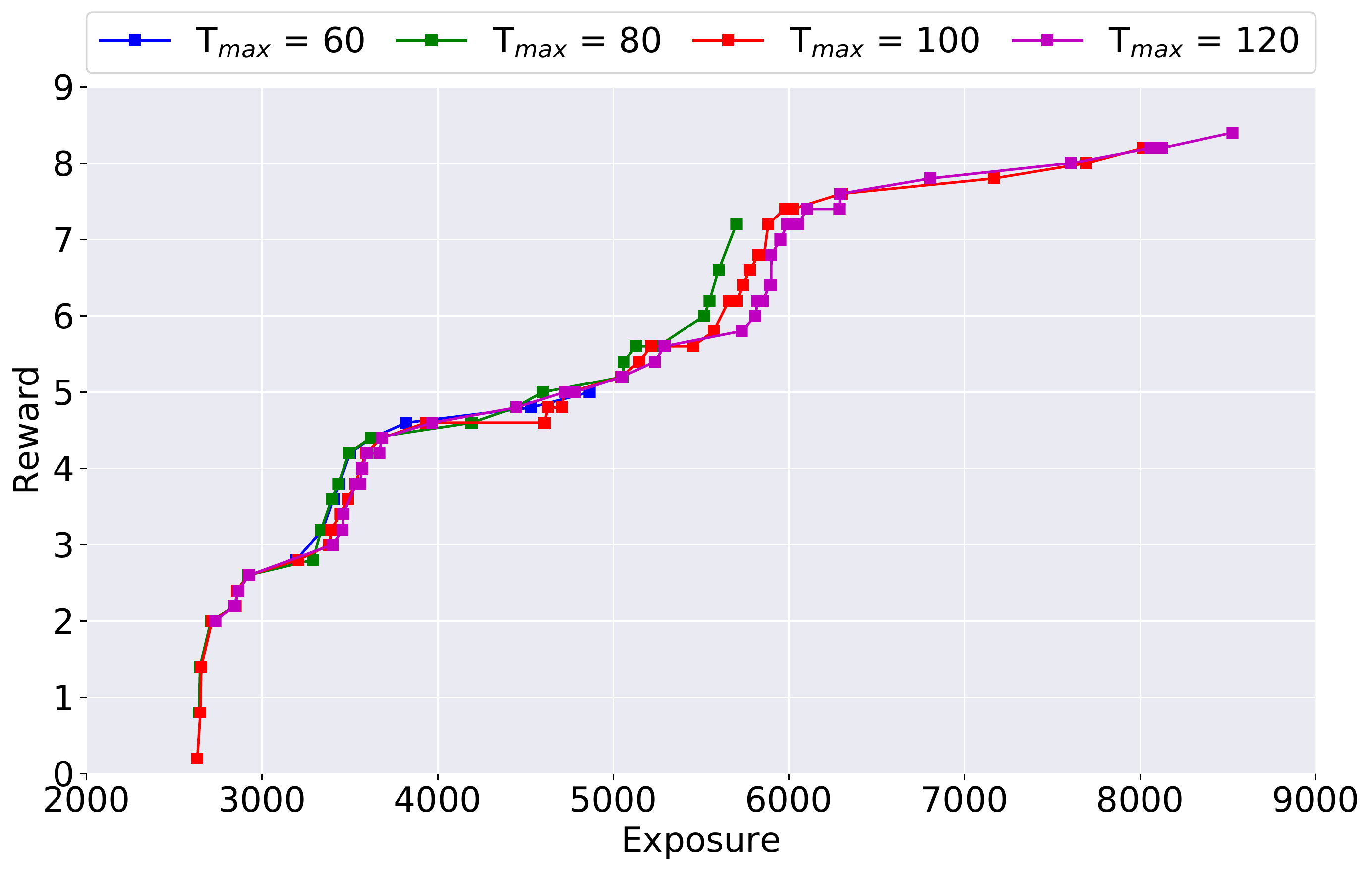}
    \caption{Test Instance - Scenario A. Final Pareto~fronts for different values of $\budget$ and fixed $\rhomax = 2.0$.}
    \label{fig:multi-pareto-fixed-rhomax}
\end{figure}

Figure \ref{fig:multi-pareto-fixed-tmax} shows the results for a fixed $\budget = 100$ and varying $\rhomax$. The following can be observed in this case: (i) it is possible to achieve a reduction on the exposure with greater values of $\rhomax$; (ii) to avoid certain regions and still respect the travel budget some locations will have to be removed from the path, resulting in a less collected reward. Hence, an intermediary value for $\rhomax$ seems the best choice.

\begin{figure}[htpb]
    \centering
    \includegraphics[width=.8\linewidth]{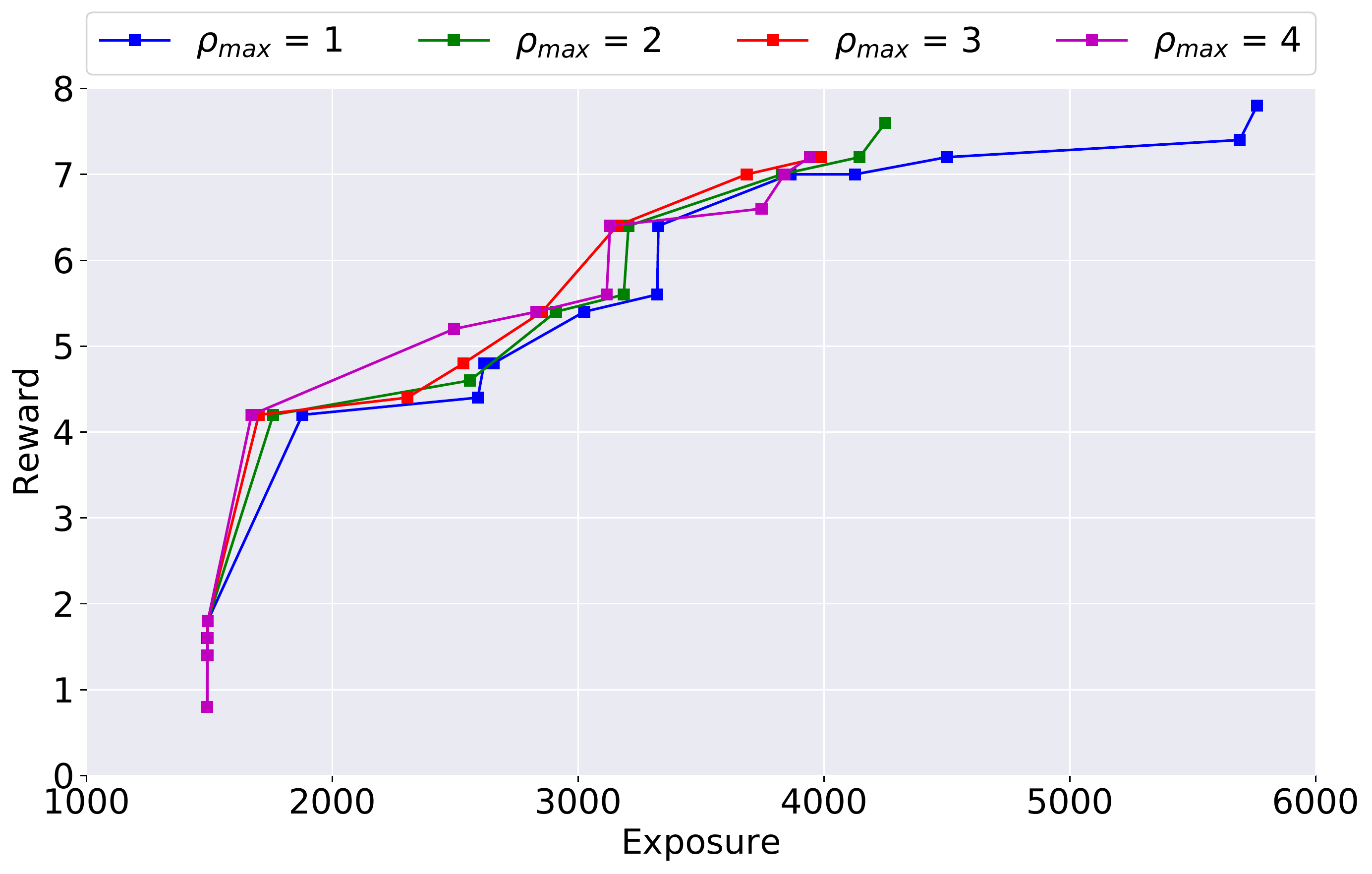}
    \caption{Test Instance - Scenario B. Final Pareto~fronts for different values of $\rhomax$ and fixed $\budget = 100$.}
    \label{fig:multi-pareto-fixed-tmax}
\end{figure}

% {
% \color{blue}

% To test the effectiveness of the designed algorithm in solving the MEP-CC problem that contains protected regions, we designed a series of simulation experiments that include scenarios with deterministic and random distributions of sensor nodes outside the protected region.

% }

% \begin{itemize}
%     \item Vary budget;
%     \item Vary rho range;
% \end{itemize}

Due to space limitation, just partial results are presented here, the complete set of results can be found online\footnote{\url{https://www.dcc.ufmg.br/~doug/medop/results.pdf}}.

\subsection{Closed path}

%In this paper, we consider a closed path ($\position{1} = \position{\nwaypoints}$), and we formulate the problem as a \ac{DTSP} instance. 

The \ac{OP} was originally called Selective Traveling Salesman Problem \cite{Laporte1990Selective}, since it considered a circuit (i.e., $\position{1} = \position{\nwaypoints}$). Such formulation is a good representation of real-world scenarios, where vehicles usually leave and return to a common location (base). Considering this case, our problem can be seen as a variant of the \ac{DTSP} with budget and exposure constraints, and the methodology can be applied directly.

Figure \ref{fig:closedpath} shows solutions of our approach for this case, with $\budget = 120$ and different intervals for the turning radius. As can be seen, with the chance of varying the turning radius it is possible to modify the path and reduce the exposure, however, with an increase in the length. This corroborates with the previous sections, showing that focusing on achieving the shortest path is not the appropriate decision for the \ac{MEDOP}. 

% - Evaluate the extremes: length, detection
% - Increase Tmax -> DTSP?
% - How does having different rhomin behave on each case?

\begin{figure}[htpb]
    \centering
    \subfloat[{$\rhomin = 1.0, \rhomax = 1.0 $}]{\includegraphics[width=.8\linewidth]{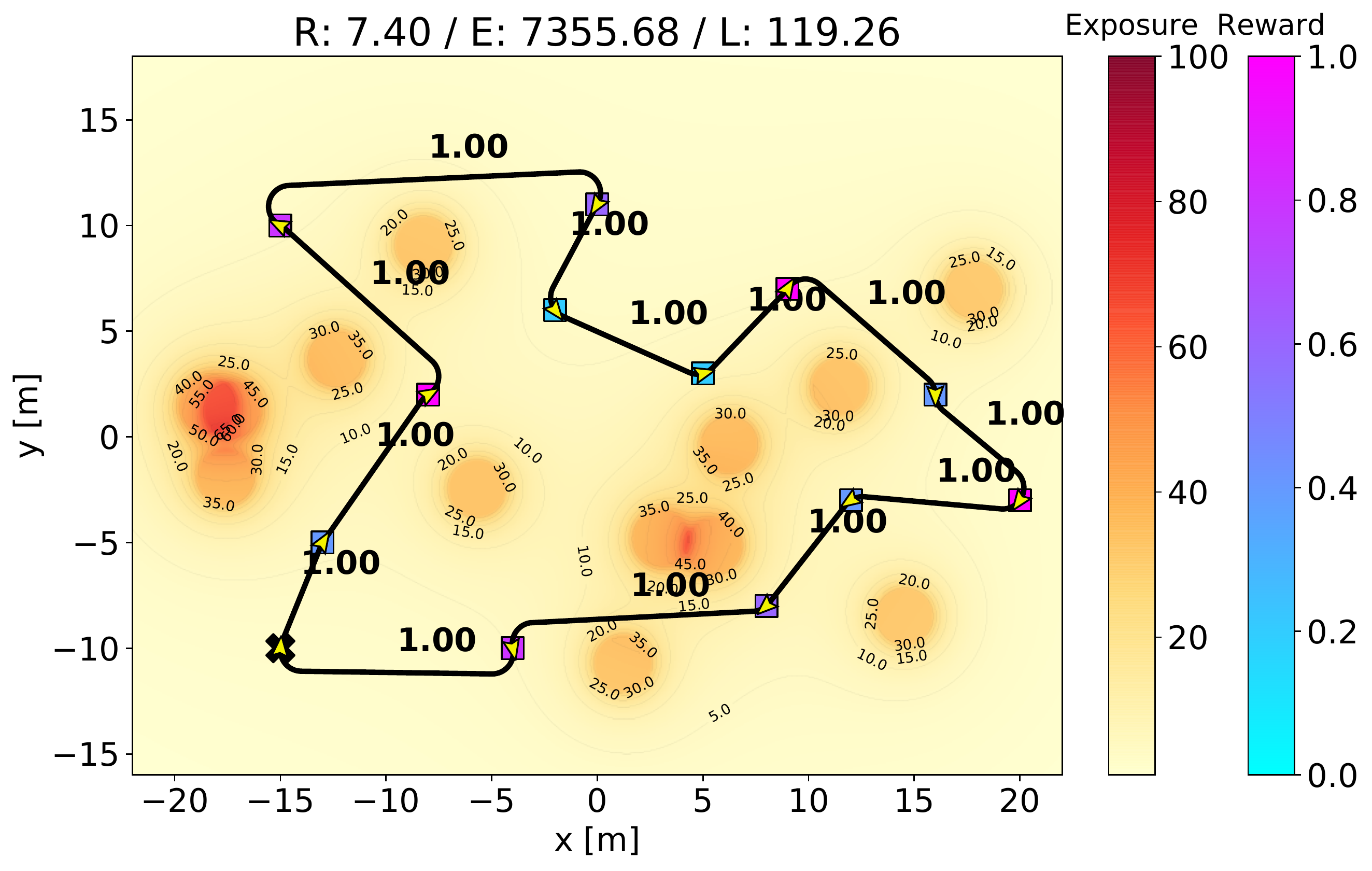}
    }
    \\
    \subfloat[{$\rhomin = 1.0, \rhomax = 4.0$}]{{\includegraphics[width=.8\linewidth]{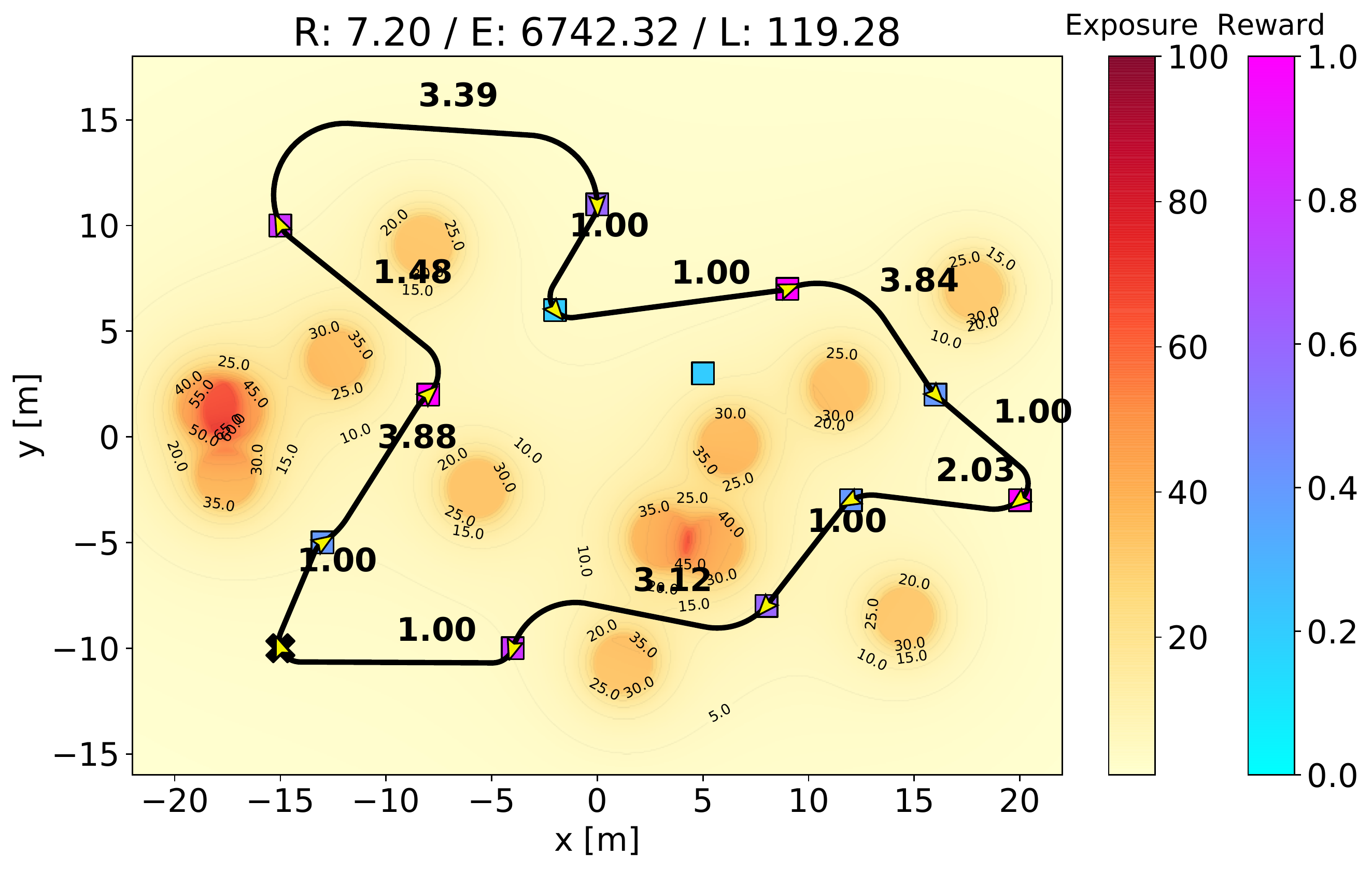}}
    }
    \caption{Closed path examples ($\budget = 120$). The bold numerals represent the turning radius used on that segment.}
    \label{fig:closedpath}
\end{figure}

%%%%%%%%%%%%%%%%%%%%%%%%%%%%%%%%%%%%%%%%%%%%%%%%%%%%%%%%%%%%%%
%%%%%%%%%%%%%%%%%%%%%%%%%%%%%%%%%%%%%%%%%%%%%%%%%%%%%%%%%%%%%%  
\section{Conclusion and Future Work}
\label{sec:conclusion}

In this paper, we introduced the \ace{MEDOP}, a multi-objective routing problem for bounded-curvature vehicles. The goal is to compute a path that maximizes the collected reward in predefined locations, while minimizes the exposure to threatening zones. This path is subjected to a limited travel budget, for example, time, length, or energy consumption.

The proposed methodology consists of an evolutionary algorithm that, in a combined manner, evaluates different sequences of visits, orientation on each location, and turning radius on each segment. Since our cost function is based on two objectives, we generate a set of fully viable solutions, and our evaluation is based on the Pareto front.

Results show that our approach can produce outcomes that have an adequate compromise between both objectives. Furthermore, it also contributes to the diversity of the solutions, which is a fundamental requirement in a decision-making step of multi-objective optimization problems.

Future research directions includes the use of multiple vehicles and more realistic environments filled with static or dynamic obstacles \cite{Penicka2019Physical}. We also intend to address heterogeneous and dynamic sensor fields (e.g., moving nodes), and time-varying rewards on the locations.

% \begin{itemize}
%     \item Multiple vehicles;
%     \item Unknown environments/with obstacles;
%     \item Online version: Time-varying profits/field.
% \end{itemize}

% {
% \color{blue}

% The all sensor intensity model reflects more accurately the capability of the sensors to detect a target, but it has also a number of weaknesses: (i) it assumes that all sensors are active during most of the time, which would be energy inefficient; (ii) it presents greater
% communication and data fusion challenges; (iii) the collection of data from weak sources increases the total noise-to-signal ratio. \cite{Djidjev2007Efficient}

% }

%%%%%%%%%%%%%%%%%%%%%%%%%%%%%%%%%%%%%%%%%%%%%%%%%%%%%%%%%%%%%%
%%%%%%%%%%%%%%%%%%%%%%%%%%%%%%%%%%%%%%%%%%%%%%%%%%%%%%%%%%%%%%
%\section*{APPENDIX}
%Appendixes should appear before the acknowledgment.

\section*{Acknowledgment}

Partially financed by the Coordenação de Aperfeiçoamento de Pessoal de Nível Superior - Brasil (CAPES) - Finance Code 001, Conselho Nacional de Desenvolvimento Científico e Tecnológico - Brasil (CNPq), and the Fundação de Amparo à Pesquisa do Estado de Minas Gerais  - Brasil (FAPEMIG).

%%%%%%%%%%%%%%%%%%%%%%%%%%%%%%%%%%%%%%%%%%%%%%%%%%%%%%%%%%%%%%
%%%%%%%%%%%%%%%%%%%%%%%%%%%%%%%%%%%%%%%%%%%%%%%%%%%%%%%%%%%%%%
\bibliographystyle{IEEEtran}
\bibliography{bibliography}

\end{document}